\RequirePackage{fix-cm}
\documentclass[twocolumn]{svjour3}          %
\usepackage[affil-it]{authblk}

\usepackage[toc,page]{appendix}

\usepackage[utf8]{inputenc} %

\usepackage[svgnames,table]{xcolor} %
\usepackage{setspace}%

\usepackage[ruled,vlined]{algorithm2e}

\usepackage[hidelinks]{hyperref}
\hypersetup{
    colorlinks,
    citecolor=blue,
    filecolor=blue,
    linkcolor=blue,
    urlcolor=blue
    }

\usepackage{graphicx} %
\usepackage{caption,subcaption} %
\usepackage{stmaryrd}
\usepackage{bbm}
\usepackage{amsmath,amsfonts,amssymb}

\usepackage{amsthm} %

\usepackage{xspace}
\usepackage{array}
\usepackage{url}
\usepackage{mathtools}
\usepackage{algorithm2e}
\usepackage{code}

\usepackage{tikz}
\usetikzlibrary{arrows,shapes,positioning}
\usetikzlibrary{decorations.markings}

\usepackage{makecell}

\usepackage{mathtools} %

\usepackage[affil-it]{authblk}
\usepackage[utf8]{inputenc} %
\usepackage[svgnames,table]{xcolor} %
\usepackage{setspace}%
\usepackage[ruled,vlined]{algorithm2e}

\usepackage{csquotes}

\usepackage{graphicx} %
\usepackage{caption,subcaption} %
\usepackage{stmaryrd}
\usepackage{bbm}
\usepackage{amsmath,amsfonts,amssymb}
\usepackage{xspace}
\usepackage{array}
\usepackage{url}
\usepackage{mathtools}
\usepackage{algorithm2e}
\usepackage{code}

\usepackage{tikz}
\usetikzlibrary{arrows,shapes,positioning}
\usetikzlibrary{decorations.markings}
\usepackage{makecell}
\usepackage{mathtools} %

\usepackage{xspace}
\usepackage{graphicx}
\usepackage{amsfonts}
\usepackage{amsmath}
\usepackage{booktabs}

\usepackage{amsthm}

\newcommand{\Morse}{$\mathfrak{D}$-Morse\xspace}
\newcommand{\connectedcomponent}{\mathcal{CC}\xspace}
\newcommand{\realscomplete}{\overline{\Reals}\xspace}
\newcommand{\iv}{\mathrm{iv}}

\newcommand{\xminall}{x^{\forall}_{\mathrm{min}}\xspace}

\newcommand{\xmindeux}{\xmin^2\xspace}

\newcommand{\xmaxdeux}{\xmax^2\xspace}
\newcommand{\xmax}{\mathbf{x}_{\max}\xspace}
\newcommand{\DYN}{\mathrm{dyn}\xspace}
\newcommand{\Proof}{\textbf{Proof:}\xspace}

\newcommand{\xsad}{\mathbf{x_{\mathrm{sad}}}\xspace}

\newcommand{\xmin}{\mathbf{x_{\mathrm{min}}}\xspace}
\newcommand{\CARD}{\mathrm{Card}\xspace}
\newcommand{\CC}{\mathcal{CC}\xspace}
\newcommand{\Reals}{\mathbb{R}\xspace}
\newcommand{\veczero}{\mathbf{0}\xspace}
\newcommand{\vecx}{\mathbf{x}\xspace}

\newcommand{\IDENTITYMATRIXNMOINSUN}{\mathbb{I}_{n-1}\xspace}
\newcommand{\XLOWER}{\mathbf{x_{<}}\xspace}
\newcommand{\ILOWER}{\mathbf{i_{<}}\xspace}
\newcommand{\DXMIN}{(D_{\xmin})\xspace}
\newcommand{\dynamics}{\mathrm{dyn}\xspace}
\newcommand{\PIOPT}{\Pi^{\mathrm{opt}}\xspace}
\newcommand{\effort}{\mathrm{Effort}\xspace}

\newcommand{\lmin}{\ell_{min}\xspace}
\newcommand{\lmax}{\ell_{max}\xspace}
\newcommand{\lstar}{\ell^*\xspace}
\newcommand{\xstar}{x^*\xspace}
\newcommand{\VOISINAGEXSTAR}{\bar{B}(\xstar,\varepsilon)\xspace}
\newcommand{\VOISINAGEZERO}{\bar{B}(\veczero,\varepsilon)\xspace}
\newcommand{\CMIN}{C^{\mathrm{min}}\xspace}
\newcommand{\imin}{i_{\mathrm{min}}\xspace}

\newcommand{\PIINF}{\Pi_<\xspace}
\newcommand{\CLOWER}{C^<\xspace}

\newcommand{\CSAD}{C^{\mathrm{sad}}\xspace}
\newcommand{\ISAD}{I^{\mathrm{sad}}\xspace}

\newcommand{\CSADLOWER}{C_{\ILOWER}^{\mathrm{sad}}\xspace}

\newcommand{\CSADI}{C_i^{\mathrm{sad}}\xspace}
\newcommand{\CSADIMIN}{C_{\imin}^{\mathrm{sad}}\xspace}
\newcommand{\CoCo}{\mathcal{CC}\xspace}
\newcommand{\CI}{C^I_i\xspace}
\newcommand{\CLO}{\mathrm{clo}\xspace}
\newcommand{\rep}{\mathrm{rep}\xspace}
\newcommand{\per}{\mathrm{Per}\xspace}

\newcommand{\BFRAK}{\mathfrak{B}\xspace}

\newcommand{\MATRICEHESSENNESADDLE}{
\begin{bmatrix}
-1 & \veczero\\
\veczero & \IDENTITYMATRIXNMOINSUN\\
\end{bmatrix}
}

\newcommand{\CINF}{\ensuremath{\mathcal{C}^{\infty}(\Reals^n)\xspace}}
\newcommand{\XMAXSAD}{\mathbf{x}_{\mathrm{max/sad}}\xspace}
\newcommand{\xmaxplus}{\xmax^+\xspace}
\newcommand{\xmaxmoins}{\xmax^-\xspace}
\newcommand{\HYPO}{\textbf{(H)}\xspace}
\newcommand{\ProofNico}{\textbf{Proof}\xspace}

\newcommand{\cloRB}{\mathrm{cl}_{\realscomplete}\xspace}
\newcommand{\ipaired}{i_{\mathrm{paired}}\xspace}

\newcommand{\CPLUS}{C_+\xspace}
\newcommand{\CMINUS}{C_-\xspace}
\newcommand{\XUNM}{x_1^M\xspace}
\newcommand{\XUNMPRIME}{x_1^{M'}\xspace}
\newcommand{\XIM}{(x_i^M)\xspace}
\newcommand{\XIMPRIME}{(x_i^{M'})\xspace} 
\newcommand{\HYP}{\mathrm{HYP}\xspace}
\newcommand{\HYPDEUX}{\mathrm{HYP2}\xspace}
\newcommand{\bd}{\partial\xspace}
\newcommand{\CIMIN}{C_{\imin}\xspace}
\newcommand{\PISTAR}{\Pi^*\xspace}

\newcommand{\nico}[1]{\textcolor{blue}{#1}\xspace}

\newtheorem{Def}{Definition}
\newtheorem{Th}{Theorem}

\newtheorem{Lem}{Lemma}
\newtheorem{Proposition}{Proposition}

\providecommand{\keywords}[1]{\textbf{\textit{Keywords: }} #1}

\begin{document}
\title{Some equivalence relation between persistent homology and morphological dynamics}
\titlerunning{Some equivalence relation between PH and MD}

\author{
Nicolas Boutry \and Laurent Najman \and Thierry G\'eraud
}
\authorrunning{N. Boutry et al.}

\institute{
Nicolas Boutry \at
EPITA Research and Development Laboratory (LRDE) \\
14-16 rue Voltaire, FR-94276 Le Kremlin-Bic\^etre, France\\
\email{nicolas.boutry@lrde.epita.fr}
\and
Laurent Najman \at
Universit\'e Gustave Eiffel, LIGM, \'Equipe A3SI, ESIEE\\
\email{laurent.najman@esiee.fr}
\and
 Thierry G\'eraud \at
EPITA Research and Development Laboratory (LRDE) \\
14-16 rue Voltaire, FR-94276 Le Kremlin-Bic\^etre, France\\
\email{thierry.geraud@lrde.epita.fr}
}

\markboth{FIX-ME1}{
\MakeLowercase{FIX-ME2}}

\date{Received: date / Accepted: date}

\maketitle              %
\begin{abstract}

In Mathematical Morphology (MM), connected filters based on dynamics are used to filter the extrema of an image. %
Similarly, persistence is a concept coming from Persistent Homology (PH) and Morse Theory (MT) that represents the stability of the extrema of a Morse function. Since these two concepts seem to be closely related, in this paper we examine their relationship, and we prove that they are equal on $n$-D Morse functions, $n\geq 1$. More exactly, pairing a minimum with a $1$-saddle by dynamics or pairing the same $1$-saddle with a minimum by persistence leads exactly to the same pairing, assuming that the critical values of the studied Morse function are unique. This result is a step further to show how much topological data analysis and mathematical morphology are related, paving the way for a more in-depth study of the relations between these two research fields.

\end{abstract}

\section{Introduction}

\begin{figure}
\centering
\includegraphics[width=\linewidth]{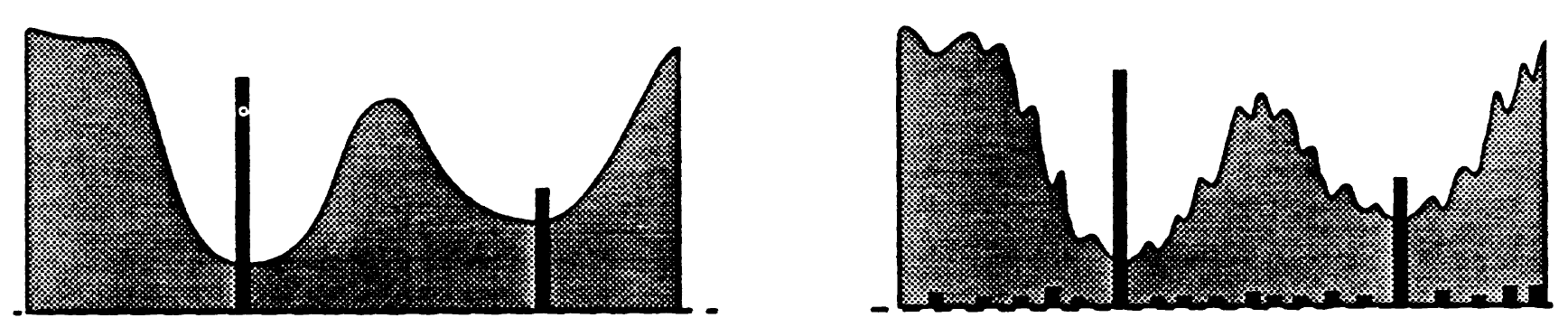}
\caption{Low sensibility of dynamics to noise (extracted from~\cite{grimaud1992new}).}
\label{fig.grimaud.2}
\end{figure}

In \emph{Mathematical Morphology}~\cite{najman2013mathematical,serra1986introduction,serra2012mathematical}, \emph{dynamics}~\cite{grimaud1991geodesie,grimaud1992new,vachier1995extraction}, defined in terms of continuous paths and optimization problems, represents a very powerful tool to measure the significance of extrema in a gray-level image (see Figure~\ref{fig.grimaud.2}). Thanks to dynamics, we can efficiently select markers of objects in an image.  These markers (that do not depend on the size or on the shape of objects) help to select relevant components  in an image; hence, this process is a way to filter objects depending on their contrast, whatever the scale of the objects, and is often combined with the watershed~\cite{najman1996geodesic,vincent1991watersheds} for image segmentation.  This contrasts with convolution filters  often used in digital signal processing or morphological filters~\cite{najman2013mathematical,serra1986introduction,serra2012mathematical} where geometrical properties do matter. 

\medskip

\begin{figure}
\centering
\includegraphics[width=\linewidth]{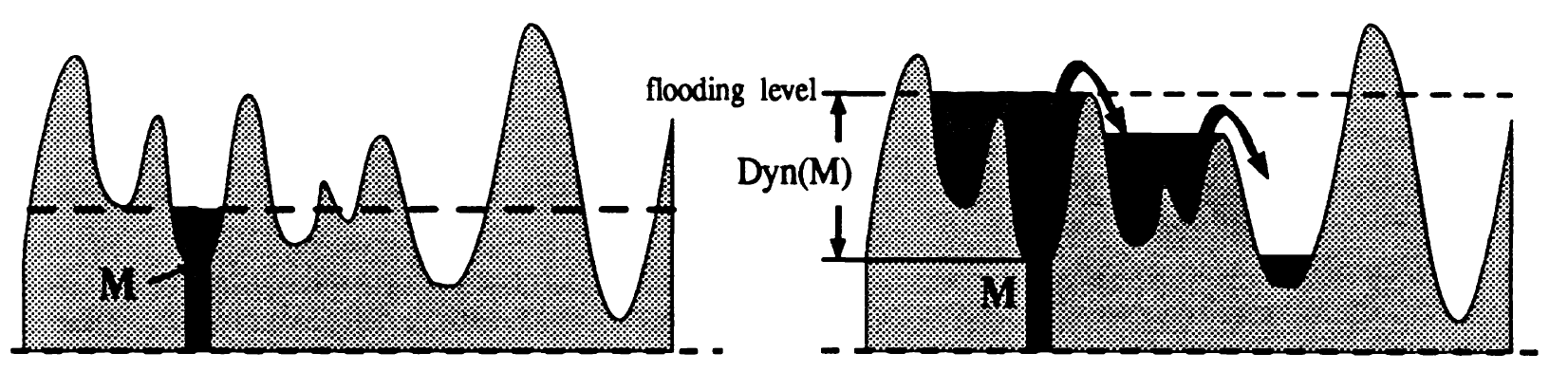}
\caption{The dynamics of a minimum of a given function can be computed thanks to a flooding algorithm (extracted from~\cite{grimaud1992new}).}
\label{fig.grimaud.3}
\end{figure}

Note that there exists an interesting relation between flooding algorithms and the computation of dynamics (see Figure~\ref{fig.grimaud.3}). Indeed, when we flood the topographical view of a function,  at a given level, two basins merge, and the dynamics of the highest minima of the two basins is the difference between the current level of water and the altitude of this highest minima.

\medskip

Similarly, in \emph{Persistent Homology}~\cite{edelsbrunner2008persistent,edelsbrunner2000topological} well-known in \emph{Computational Topology}~\cite{edelsbrunner2010computational}, we can find the same paradigm: topological features whose~\emph{persistence} is high are \textquote{true} when the ones whose persistence is low are considered as sampling artifacts, whatever their scale. An example of application of persistence is the filtering of \emph{Morse-Smale complexes}~\cite{edelsbrunner2003hierarchical,edelsbrunner2003morse,gunther2012efficient} used in \emph{Morse Theory}~\cite{milnor1963morse,forman2002user} where pairs of extrema of low persistence are canceled for simplification purpose. This way, we obtain simplified topological representations of \emph{Morse functions}. A discrete counterpart of Morse theory, known as \emph{Discrete Morse Theory} can be found in~\cite{forman1995discrete,jollenbeck2009minimal,forman2002user,forman1998morse}.

\medskip

As detailed in~\cite{dey2007stability}, pairing by persistence of critical values can be extended in a more general setting to pairing by \emph{interval persistence} of critical points. The result is that it is possible to perform function matching based on their critical points, and then  to pair all critical points of a given function (see Figure~2 in~\cite{dey2007stability}) where persistent homology does not succeed. However, due to the modification of the definition  introduced in~\cite{dey2007stability}, this matching is not applicable when we consider usual threshold sets.

\medskip

In this paper, we prove that the relation between Mathematical Morphology and Persistent Homology is strong in the sense that pairing (of minima) by dynamics and pairing $1$-saddles by persistence is equivalent (and then dynamics and persistence of the corresponding pair are equal) in $n$-D ($n\geq 1$), when we work with Morse functions. For $n=1$, the proof is much simpler (with some extra condition on the limits of the domain), but contains the essence of the proof for $n\geq 1$, which is more technical. In order to ease the reading, we provide the complete proofs for both cases, first for the 1D case and then for the $n$-D case. This paper is the extension of \cite{boutry2019equivalence} (which contains the 1D case) and \cite{boutry2021equivalence} (which generalizes \cite{boutry2019equivalence} to the $n$-D case, $n\geq 1$).

\medskip

The plan of the paper is the following: Section~\ref{sec.background} recalls the mathematical background needed in this paper, Section~\ref{sec.sketches} provides sketches of the equivalence of pairing by dynamics and by persistence in 1D and in $n$-D, Section~\ref{sec.equivalence1D} contains the complete proof of the 1D equivalence, while Section~\ref{sec.equivalence} contains the complete proof of the $n$-D equivalence. In Section~\ref{sec:perspectives}, we discuss several research directions opened by the results of this paper. Section~\ref{sec.conclusion} concludes the paper.

\section{Mathematical pre-requisites}
\label{sec.background}

We call \emph{path} from $\vecx$ to $\vecx'$ both in $\Reals^n$ a continuous mapping from $[0,1]$ to $\Reals^n$. Let $\Pi_1$, $\Pi_2$ be two paths satisfying $\Pi_1(1) = \Pi_2(0)$, then we denote by $\Pi_1 <> \Pi_2$ the \emph{join} between these two paths. For any two points $\vecx^1,\vecx^2 \in \Reals^n$, we denote by $[\vecx^1,\vecx^2]$ the path:
$$\lambda \in [0,1] \rightarrow (1-\lambda) . \vecx^1 + \lambda . \vecx^2.$$

\medskip

Also, we work with $\Reals^n$ supplied with the Euclidean norm: $$\|.\|_2 : \vecx \rightarrow \|\vecx\|_2 = \sqrt{\sum_{i = 1}^n \vecx_i^2}.$$

\medskip

In the sequel, we use \emph{lower threshold sets} coming from cross-section topology~\cite{meyer1989skeletons,bertrand1996topological,beucher1992morphological} of a function $f$ defined for some real value $\lambda \in \Reals$ by:
$$[f < \lambda] = \left\{x \in \Reals^n \ \Big| \ f(x) < \lambda \right\},$$
and
$$[f \leq \lambda] = \left\{x \in \Reals^n \ \Big| \ f(x) \leq \lambda \right\}.$$

\subsection{Morse functions}

We call \emph{Morse functions} the real functions in $\CINF$ whose Hessian is not degenerated at \emph{critical values}, that is, where their gradient vanishes. A strong property of Morse functions is that their critical values are isolated. In particular, we call \Morse functions the Morse functions which tend to $\pm\infty$ when the $2$-norm of their argument tends to $+\infty$. Note that this last property will only be used to treat the 1D case in this paper.

\begin{Lem}[Morse Lemma~\cite{audin2014morse}]
\label{lemma.morse}
Let $f : \CINF \rightarrow \Reals$ be a Morse function. When $\xstar \in \Reals^n$ is a critical point of $f$, then there exists some neighborhood $V$ of $\xstar$ and some diffeomorphism $\varphi : V \rightarrow \Reals^n$ such that $f$ is equal to a second order polynomial function of $\vecx = (x_1,\dots,x_n)$ on $V$: $\forall \; \vecx \in V$, $$f \circ \varphi^{-1} (\vecx) = f(\xstar) - x_1^2 - x_2^2 - \dots - x_k^2 + x_{k+1}^2 + \dots + x_n^2.$$

\end{Lem}

We call \emph{$k$-saddle} of a Morse function a point $x \in \Reals^n$ such that the Hessian matrix has exactly $k$ strictly negative eigenvalues (and then $(n-k)$ strictly positive eigenvalues); in this case, $k$ is sometimes called the \emph{index} of $f$ at $x$. We say that a Morse function has \emph{unique critical values} when for any two different critical values $x_1,x_2 \in \Reals^n$ of $f$, we have $f(x_1) \neq f(x_2)$. (See Appendix \ref{app:ambiguities} for a discussion \nico{about} this hypothesis.)

\subsection{Pairing by dynamics (1D)}
\label{ssec.dyn}

\begin{figure}
\centering
\includegraphics[width=0.6\linewidth]{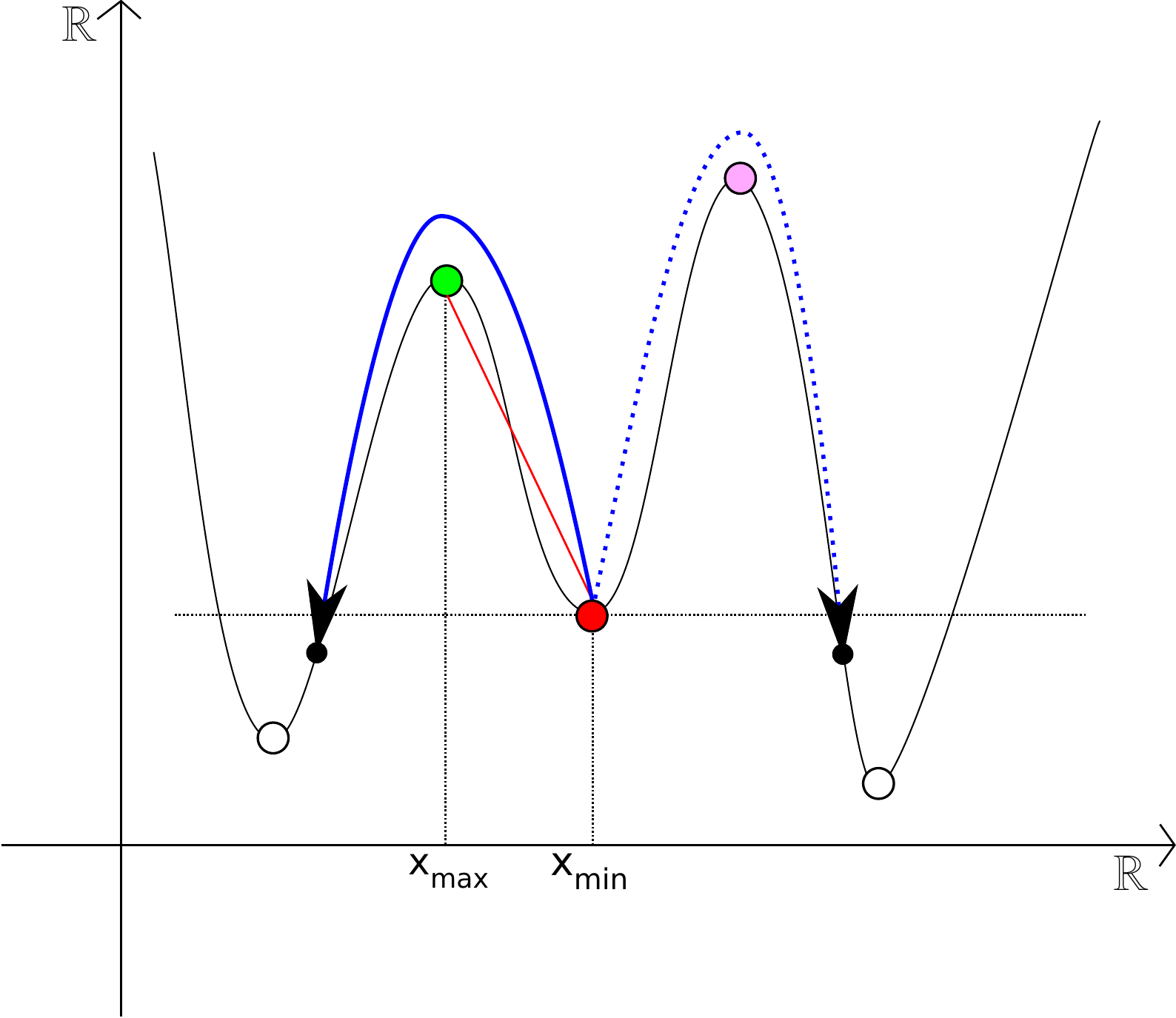}
\caption{Example of pairing by dynamics: the abscissa $\xmin$ of the red point is paired by dynamics relatively to $f$ with the abscissa $\xmax$ of the green point on its left because the \textquote{effort} needed to reach a point of lower height than $f(\xmin)$ (like the two black points) following the graph of $f$ is minimal on the left.
}
\label{fig.pairingbydynamics}
\end{figure}

Let $f : \Reals \rightarrow \Reals$ be a \Morse function with unique critical values. For $\xmin \in \Reals$ a local minimum of $f$, if there exists at least one abscissa $\xmin' \in \Reals$ of $f$ such that $f(\xmin') < f(\xmin)$, then we define the \emph{dynamics}~\cite{grimaud1992new} of $\xmin$ by:

$$\DYN(\xmin) := \min_{\gamma \in C} \max_{s \in [0,1]} f(\gamma(s)) - f(\xmin),$$
\noindent
where $C$ is the set of paths $\gamma : [0,1] \rightarrow \Reals$ verifying $\gamma(0) := \xmin$ and verifying that there exists some $s \in ]0,1]$ such that $f(\gamma(s)) < f(\xmin)$.

\medskip

Let us now define $\gamma^*$ as a path of $C$ verifying: 
$$\max_{s \in [0,1]} f(\gamma^*(s)) - f(\xmin) = \min_{\gamma \in C} \max_{s \in [0,1]} f(\gamma(s)) - f(\xmin),$$
then we say that this path is \emph{optimal}. The real value $\xmax$ \emph{paired by dynamics} to $\xmin$ (relatively to $f$) is the local maximum of $f$ characterized by:
$$\xmax := \gamma^*(s^*),$$
with $f(\gamma^*(s^*)) = \max_{s \in [0,1]} f(\gamma^*(s))$. We obtain then:
$$f(\xmax) - f(\xmin) = \DYN(\xmin).$$

Note that the local maximum $\xmax$ of $f$ does not depend on the path $\gamma^*$ (see Figure~\ref{fig.pairingbydynamics}), and its value is unique (by hypothesis on $f$), which shows that in some way $\xmax$ and $\xmin$ are \textquote{naturally} paired by dynamics.

\subsection{Pairing by persistence (1D)}

From now on, we denote by $\realscomplete := \{+\infty,-\infty\} \cup \Reals$ the complete real line, and by $\cloRB(A)$ the closure in $\realscomplete$ of the set $A \subseteq \Reals$.

\medskip

\begin{algorithm}[h]
\caption{Pairing by persistence of $\xmax$.}
\label{algo.pairingbypersistence}
$\xmin := \emptyset$\;
$[\xmax^-,\xmax^+] := \cloRB (\connectedcomponent([f \leq f(\xmax)],\xmax))$\;
\If{$\xmax^- > -\infty \; \| \; \xmax^+ < +\infty$}{
$\xmin^- := \rep([\xmax^-,\xmax],f)$\;
$\xmin^+ := \rep([\xmax,\xmax^+],f)$\;
\If{$\xmax^- > -\infty \; \& \& \; \xmax^+ < +\infty$}{
$\xmin := {\arg \max}_{x \in \{\xmin^-,\xmin^+\}} f(x)$\;
}
\If{$\xmax^- > -\infty \; \&\& \; \xmax^+ = +\infty$}{
$\xmin := \xmin^-$\;
}
\If{$\xmax^- = -\infty \; \&\& \; \xmax^+ < +\infty$}{
$\xmin := \xmin^+$\;
}
}
\return $\xmin$\;
\end{algorithm}

\begin{figure}
\centering
\includegraphics[width=0.6\linewidth]{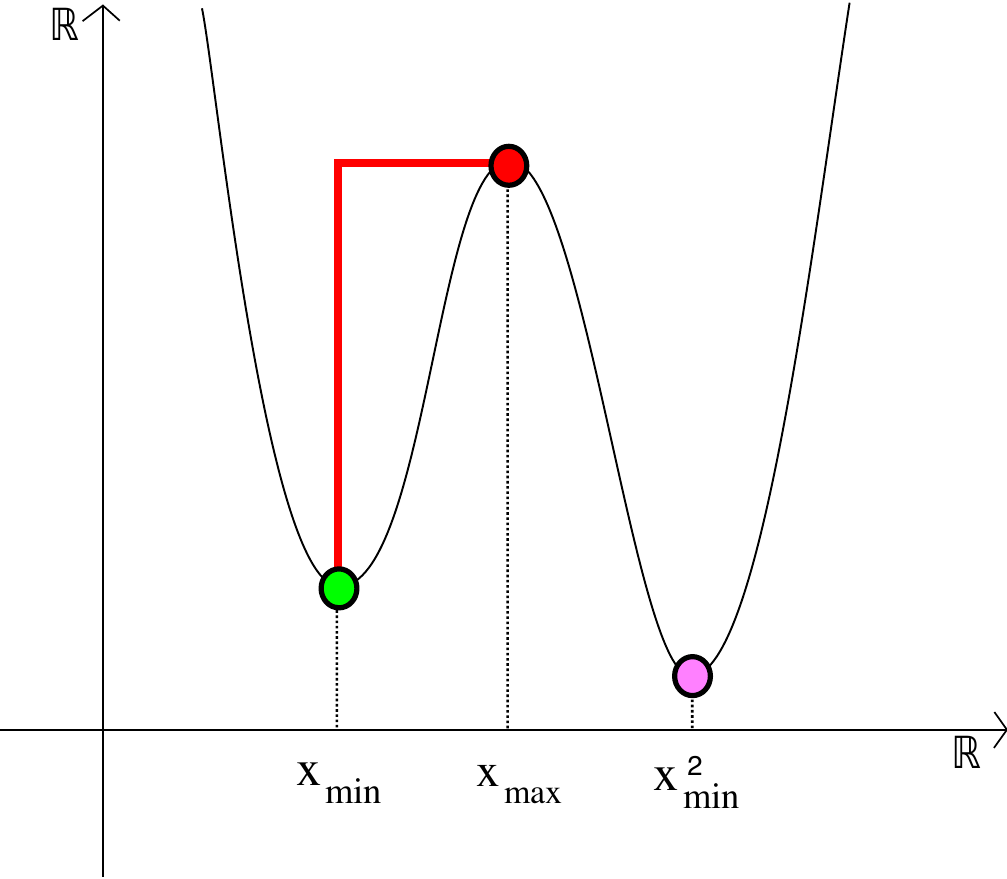}
\caption{Example of pairing by persistence: the abscissa $\xmax$ of the local maximum in red is paired by persistence relatively to $f$ with the abscissa $\xmin$ of the local minimum in green, since its image by $f$ is greater than the image by $f$ of the abscissa $\xmindeux$ of the local minimum drawn in pink.
}
\label{fig.persistence}
\end{figure}

Let $f : \Reals \rightarrow \Reals$ be a \Morse function with unique critical values, and let $\xmax$ be a local maximum of $f$. Let us recall the 1D procedure~\cite{edelsbrunner2008persistent} which pairs (relatively to $f$) local maxima to local minima (see Algorithm~\ref{algo.pairingbypersistence}). Roughly speaking, the representatives $\xmin^-$ and $\xmin^+$ are the abscissas where connected components of respectively $$[f \leq (f(\xmin^-)]\text { and }[f \leq (f(\xmin^+)]$$ \textquote{emer\-ge} (see Figure~\ref{fig.persistence}), when $\xmax$ is the abscissa where two connected components of $[f < f(\xmax)]$ \textquote{mer\-ge} into a single component of $[f \leq f(\xmax)]$. Pairing by persistence associates then $\xmax$ to the value $\xmin$ belonging to $\{\xmin^-,\xmin^+\}$ which maximizes $f(\xmin)$. The \emph{persistence} of $\xmax$ relatively to $f$ is then equal to $\per(\xmax) := f(\xmax) - f(\xmin)$.

\subsection{Pairing by dynamics ($n$-D)}

\begin{figure}
\centering
\includegraphics[width=0.7\linewidth]{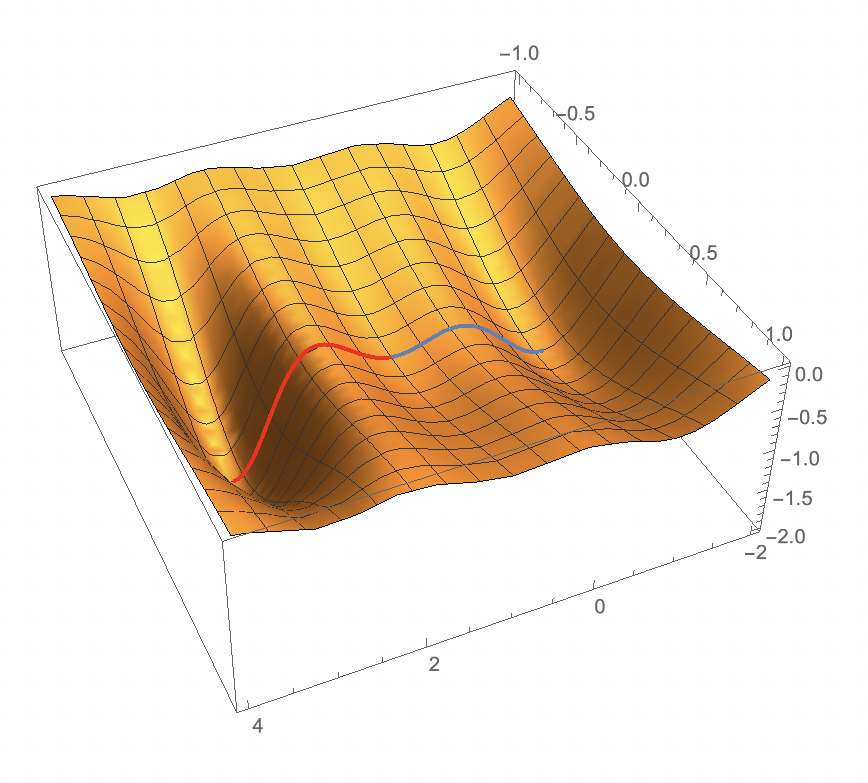}
\caption{Pairing by dynamics on a Morse function: the red and blue paths are both in $\DXMIN$ but only the blue one reaches a point $\XLOWER$ whose height is lower than $f(\xmin)$ with a minimal effort.}
\label{fig.dynpair}
\end{figure}

From now on, $f : \Reals^n \rightarrow \Reals$ is a Morse function with unique critical values.

\medskip

Let $\xmin$ be a local minimum of $f$. Then we call \emph{set of descending paths starting from $\xmin$} (shortly $\DXMIN$) the set of paths going from $\xmin$ to some element $\XLOWER \in \Reals^n$ satisfying $f(\XLOWER) < f(\xmin)$.

\medskip

The \emph{effort} of a path $\Pi : [0,1] \rightarrow \Reals^n$ (relatively to $f$) is equal to:
$$\effort(\Pi) := \max_{\ell \in [0,1], \ell' \in [0,1]} (f(\Pi(\ell)) - f(\Pi(\ell'))).$$

\medskip

A local minimum $\xmin$ of $f$ is said to be \emph{matchable} if there exists some $\XLOWER \in \Reals^n$ such that $f(\XLOWER) < f(\xmin)$. We call \emph{dynamics} of a matchable local minimum $\xmin$ of $f$ the value:
$$\dynamics(\xmin) = \min_{\Pi \in \DXMIN} \max_{\ell \in [0,1]} \left(f(\Pi(\ell)) - f(\xmin) \right),$$ 
and we say that $\xmin$ is \emph{paired by dynamics} (see Figure~\ref{fig.dynpair}) with some $1$-saddle $\xsad \in \Reals^n$ of $f$ when:
$$\dynamics(\xmin) = f(\xsad) - f(\xmin).$$

\medskip

An \emph{optimal} path $\PIOPT$ is an element of $\DXMIN$ whose effort is equal to $\min_{\Pi \in \DXMIN}(\effort(\Pi))$. Note that for any local minimum $\xmin$ of $f$, there always exists some optimal path $\PIOPT$ such that: $$\effort(\PIOPT) = \dynamics(\xmin).$$

\medskip

Thanks to the uniqueness of critical values of $f$, there exists only one critical point of $f$ which can be paired with $\xmin$ by dynamics. 

\medskip

Dynamics are always positive, and the dynamics of an absolute minimum of $f$ is set at $+\infty$ (by convention).

\subsection{Pairing by persistence ($n$-D)}

\begin{figure}
\centering
\includegraphics[width=0.45\linewidth]{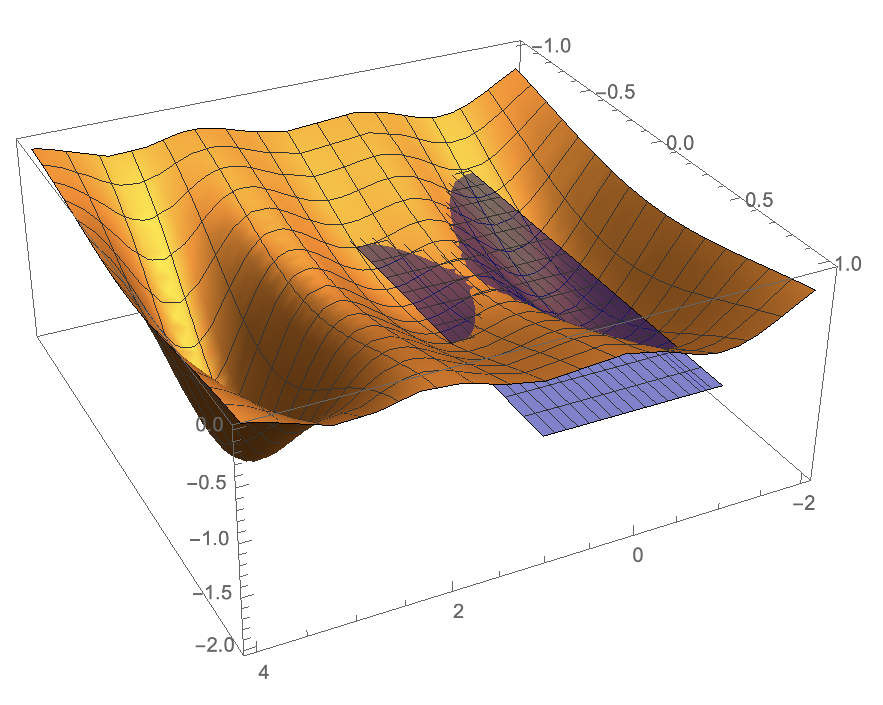}~
\includegraphics[width=0.45\linewidth]{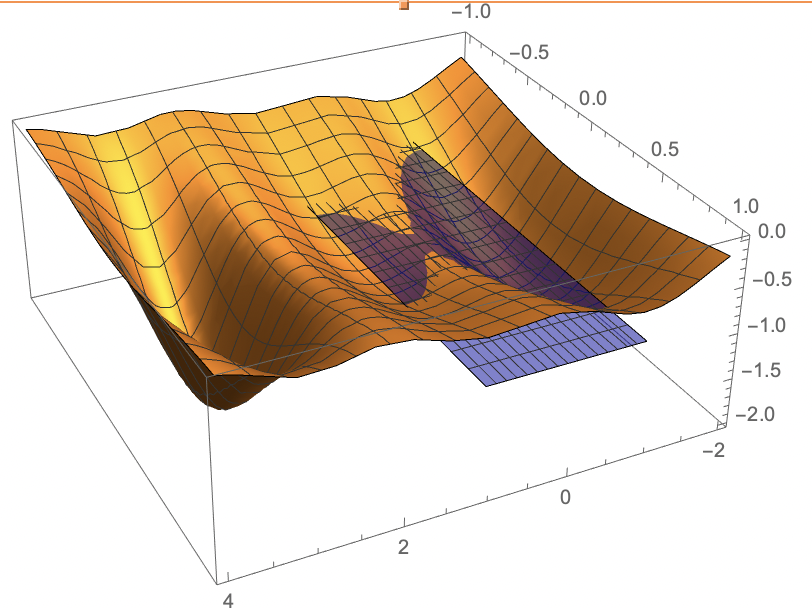}
\caption{Pairing by persistence on a Morse function: we compute the plane whose height is reaching $f(\xsad)$ (see the left side), which allows us to compute $\CSAD$, to deduce the components $\CI$ whose closure contains $\xsad$, and to decide which representative is paired with $\xsad$ by persistence by choosing the one whose height is the greatest. We can also observe (see the right side) the \emph{merge phase} where the two components merge and where the component whose representative is paired with $\xsad$ dies.}
\label{fig.perpair}
\end{figure}

Let us denote by $\CLO$ the closure operator, which adds to a subset of $\Reals^n$ all its accumulation points, and by $\CoCo(X)$ the connected components of a subset $X$ of $\Reals^n$. We also define the \emph{representative} of a subset $X$ of $\Reals^n$ relatively to a Morse function $f$ the point which minimizes $f$ on $X$:

$$\rep(X) = {\arg \min}_{\vecx \in X} f(\vecx).$$

\medskip

\begin{Def}
\label{def.persistence}
Let $f$ be some Morse function with unique critical values, and let $\xsad$ be the abscissa of some $1$-saddle point of $f$. Now we define the following expressions. First, $$\CSAD = \CC([f \leq f(\xsad)],\xsad)$$ denotes the component of the set $[f \leq f(\xsad)]$ which contains $\xsad$. Second, we denote by:
$$\{\CI\}_{i \in I} = \CoCo([f < f(\xsad)])$$
the connected components of the open set $[f < f(\xsad)]$. Third, we define 
$$\{\CSADI\}_{i \in \ISAD} = \left\{ \CI \ |\ \xsad \in \CLO(\CI) \right\}$$
the subset of components $\CI$ whose closure contains $\xsad$. Fourth, for each $i \in \ISAD$, we denote $$\rep_i = {\arg \min}_{x \in \CSADI} f(x)$$ the representative of $\CSADI$. Fifth, we define the abscissa
$$\xmin = \rep_{\ipaired}$$
with
$$\ipaired = {\arg \max}_{i \in \ISAD} f(\rep_i),$$
thus $\xmin$ is the representative of the component $\CSADI$ of minimal depth. In this context, we say that $\xsad$ is \emph{paired by persistence} to $\xmin$. Then, the \emph{persistence} of $\xsad$ is equal to:
$$\per(\xsad) = f(\xsad) - f(\xmin).$$
\end{Def}

\section{Sketches of the proofs (1D vs. $n$-D)}
\label{sec.sketches}

\subsection{Pairing by dynamics implies pairing by persistence}

\begin{table*}[!htbp]
\centering
\caption{Sketches of the 1D/$n$-D proofs that pairing by dynamics implies pairing by persistence.}
\begin{tabular}{l|c|l}
\toprule
\multicolumn{3}{c}{Hypotheses:}\\
\toprule
\toprule
$f$ is a \Morse function & & $f$ is a Morse function\\
\midrule
\multicolumn{3}{c}{$f$ has unique critical values}\\
\midrule
\multicolumn{3}{c}{$\xmin$ is a local minimum of $f$}\\
\midrule
\multicolumn{3}{c}{$\xmin$ and $\XMAXSAD$ are paired by \textbf{dynamics}}\\
\midrule
$\xmax > \xmin$ & \hspace{1em} & $\xmin \neq \xsad$\\
\toprule
\toprule
\multicolumn{3}{c}{Notations}\\
\toprule
\toprule
$[\xmaxmoins,\xmaxplus] = \cloRB (\connectedcomponent([f \leq f(\xmax)],\xmax))$ & & $\CSAD = \CC([f \leq f(\xsad)],\xsad)$\\
\midrule
& & $\{\CI\}_{i \in I} = \CoCo([f < f(\xsad)])$\\
\midrule
& & $\{\CSADI\}_{i \in \ISAD} = \left\{ \CI \ |\ \xsad \in \CLO(\CI) \right\}$\\
\bottomrule
\bottomrule
\multicolumn{3}{c}{Step 1:}\\
\midrule
& & $\exists\ i \in I$ s.t. $\xmin \in \CI$\\
$\xmin$ represents $[\xmaxmoins,\xmax]$ & & with $\xmin$ representing $\CI$\\
\midrule
\multicolumn{3}{c}{(otherwise $\DYN(\xmin) < f(\XMAXSAD) - f(\xmin)$ which leads to a contradiction)}\\
\midrule
& & $\CI$ belongs to $\{\CSADI\}_{i \in \ISAD}$\\
& & then $\xmin$ represents some $\CSADIMIN$\\
\bottomrule
\bottomrule
\multicolumn{3}{c}{Step 2:}\\
\toprule
\toprule
$f(\rep([\xmax,\xmaxplus],f)) < f(\xmin)$ & & $\forall \ i \neq \imin$, $f(\rep(\CSADI,f) < f(\xmin)$\\
\midrule
\multicolumn{3}{c}{(otherwise $\DYN(\xmin) > f(\XMAXSAD) - f(\xmin)$ which leads to a contradiction)}\\
\bottomrule
\bottomrule
\multicolumn{3}{c}{Step 3:}\\
\toprule
\toprule
\multicolumn{3}{c}{$\xmin$ and $\XMAXSAD$ are paired by persistence}\\
\bottomrule
\end{tabular}
\label{table.dyn.to.per}
\end{table*}

\begin{figure}
    \centering
    \includegraphics[width=0.7\linewidth]{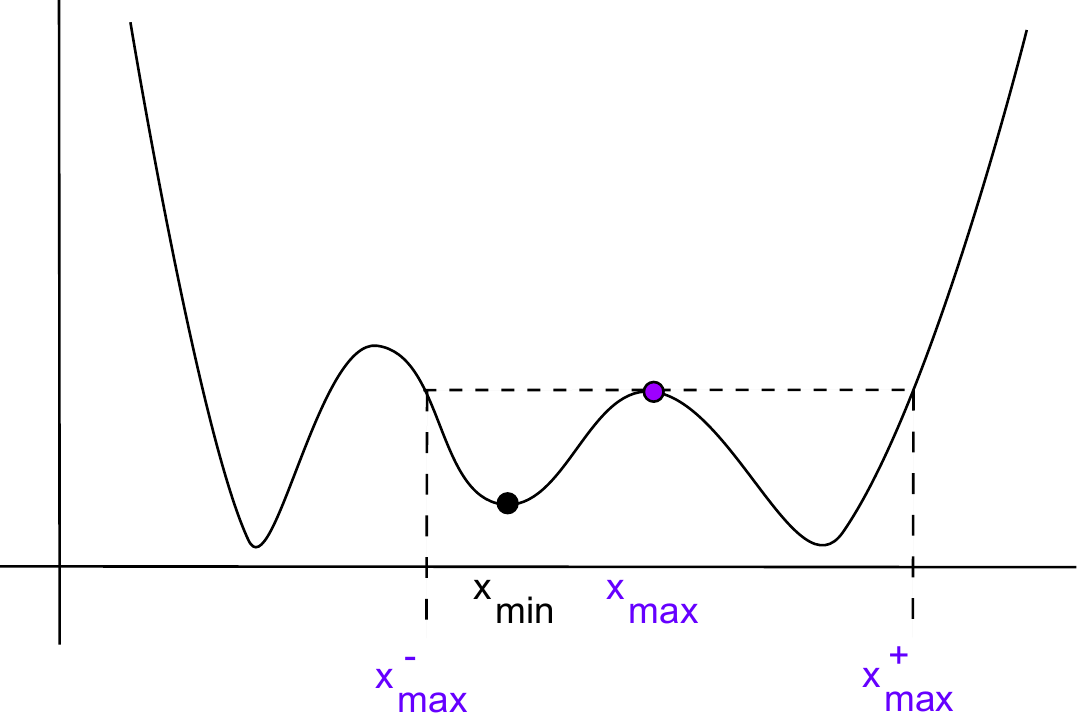}
    \caption{Pairing by dynamics implies pairing by persistence in 1D: when $\xmin$ (in black) is paired with $\xmax$ (in purple) by dynamics, we observe easily that $\xmin$ is the representative of the basin where it lies. Furthermore, the optimal path descending lower than $f(\xmin)$  goes on the right side and  goes through $\xmax$ (since we look for a minimal effort and $f(\xmaxdeux)$ is greater than $f(\xmax)$). This implies that the right basin contains a representative lower than $f(\xmin)$. Since $\connectedcomponent([f \leq f(\xmax)],\xmax)$ is made of the two described basins, we obtain easily that $\xmax$ is paired with $\xmin$ by persistence.}
    \label{fig:sketch1D-dyn-per}
\end{figure}

Let us start from the 1D case (see Figure~\ref{fig:sketch1D-dyn-per}). We assume  (see Table~\ref{table.dyn.to.per}) that we have some Morse function $f$ defined on the real line and that the critical values are unique, that is, for two different extrema $x_1,x_2$ of $f$, we have $f(x_1) \neq f(x_2)$. Furthermore, we assume that the abscissas $\{\xmin,\xmax\}$ with $\xmax > \xmin$ are paired by dynamics, that is, starting from $\xmin$ and following the graph of $f$, the lower effort to reach a lower value is on the right side. Using these properties, we want to show that $\xmax$ and $\xmin$ are paired by persistence.

\medskip

\textbf{\underline{1D proof:}} Let us proceed in three steps. First, we want to show that $\xmin$ is the representative of the basin $[\xmaxmoins,\xmax]$ of level $f(\xmax)$ containing it. This is easily proven by contradiction: if $\xmin$ is not the representative of this basin, there exists some $\xstar$ in it where $f(\xstar) < f(\xmin)$, and then the dynamics of $\xmin$ is lower than $f(\xmax) - f(\xmin)$, which is impossible by hypothesis.

\medskip

Now that we know that $\xmin$ represents the basin $[\xmaxmoins,\xmax]$, we can show that $f(\xmin)$ is greater than the image by $f$ of the representative of $[\xmax,\xmaxplus]$ corresponding also to the lower threshold set $[f \leq f(\xmax)]$. By assuming the contrary, we would imply that any descending path starting from $\xmin$ would go outside the component $[\xmaxmoins,\xmaxplus] = \connectedcomponent([f \leq f(\xmax),\xmax])$, which means that we would obtain a dynamics of $\xmin$ greater than $f(\xmax) - f(\xmin)$, which is impossible.

\medskip

Since we have obtained that $\xmin$ is the representative of the highest basin starting for the extrema $\xmax$, we can conclude easily that $\xmax$ is paired with $\xmin$ by persistence.

\medskip

\textbf{\underline{$n$-D proof:}} The proof in $n$-D, $n \geq 2$, is very similar, except that we have more complex notations. Indeed, we  study $1$-saddles instead of maxima; the path between the two points is not \textquote{unique} anymore; and we do not have anymore a natural order between two abscissas.

\medskip

We cannot define $\xmaxmoins$ and $\xmaxplus$, but instead we can define the closed connected component $\connectedcomponent([f \leq f(\xmax)],\xmax)$ containing $\xmax$. Also, we cannot define $]\xmaxmoins,\xmax[$ or $]\xmax,\xmaxplus[$  but instead we can define the connected components $\CI$ which are components of $[f < f(\xsad)]$, and the components $\CSADI$ of $[f < f(\xsad)]$ with the additional property that their closure contains $\xsad$. Last point, we do not need anymore the condition that the studied function tends to infinity when the norm of the abscissa tends to infinity, but the consequence is that the proof is a little more complex.

\medskip

After having introduced these notations, we can follow the same three steps as before. We first prove that $\xmin$, paired to $\xsad$ by dynamics, is the representative of some $\CI$ (otherwise we would obtain that the dynamics of $\xmin$ is lower than $f(\xsad) - f(\xmin)$ since we can reach a point on the graph of $f$ which is lower than $f(\xmin)$). Then, the proof that this $\CI$ is in fact one of the $\CSADI$ follows from the fact that otherwise, any descending path of $\xmin$ must go out of $\CI$ to reach a lower value than $f(\xmin)$, and then the dynamics of $\xmin$ would be greater than $f(\xsad) - f(\xmin)$.

\medskip

Now that we know that $\xmin$ belongs to some $\CSADI$, we can use the property that there exists exactly two basins in the component $\connectedcomponent([f \leq f(\xsad)],\xsad)$ (since we work with a Morse function). By assuming that $\xmin$ is not the highest representative among the open components $\CSADI$, we obtain one more time that any path starting from $\xmin$ must go outside $$\connectedcomponent([f \leq f(\xsad)],\xsad)$$ to descend lower than $f(\xmin)$, which would lead to a greater dynamics than $f(\xsad) - f(\xmin)$. Thus, $\xmin$ is the highest representative among the ones of the components $\{\CSADI\}_i$.

\medskip

We conclude one more time that $\xsad$ is paired to $\xmin$ by persistence when $\xmin$ is paired to $\xsad$ by dynamics.

\subsection{Pairing by persistence implies pairing by dynamics}

\begin{figure}
    \centering
    \includegraphics[width=0.7\linewidth]{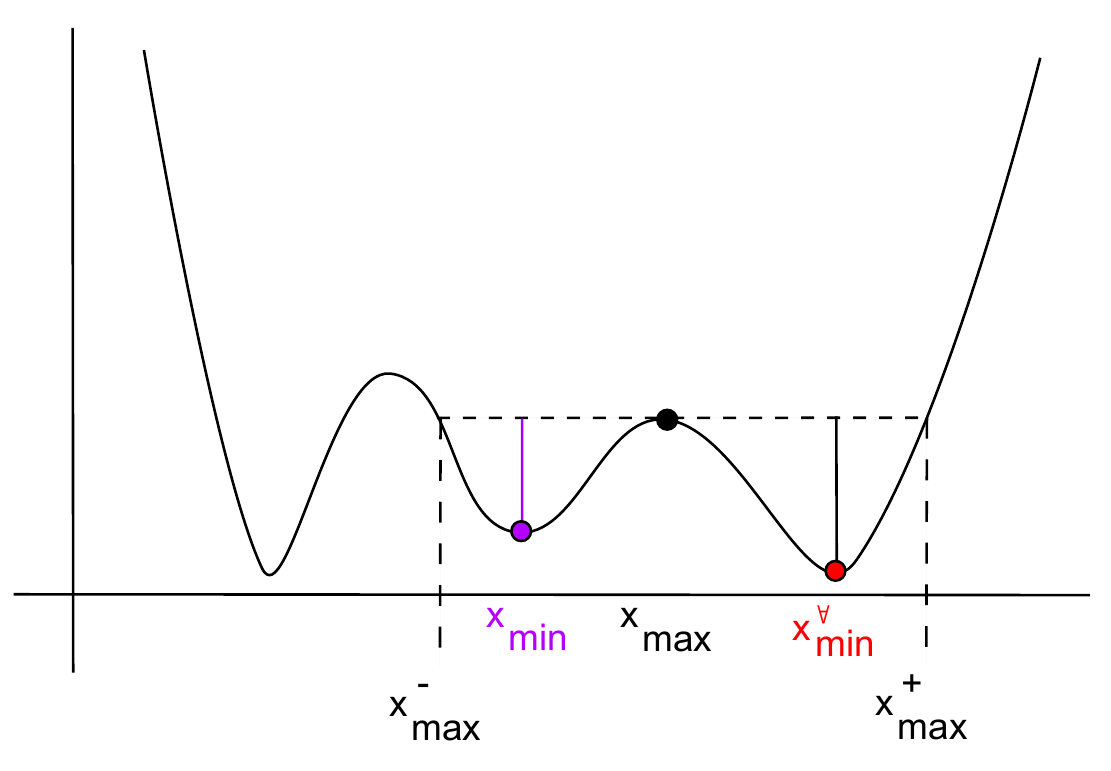}
    \caption{Pairing by persistence implies pairing by dynamics in 1D: starting from the local maximum $\xmax$ (in black), we define the component $[\xmaxmoins,\xmaxplus]$ of the lower threshold set of $f$ which contains $\xmax$. By definition of pairing by persistence, we know that the representative of the component $[\xmaxmoins,\xmax]$ is $\xmin$ drawn in purple (since $\xmin < \xmax$) and we call $\nico{\xminall}$ (drawn in red) the representative of the component $[\xmax,\xmaxplus]$. From these facts, we deduce easily that $\xmin$ is matchable since $f(\nico{\xminall}) < f(\xmin)$. We also deduce that there exists a descending path from $\xmin$ to $\xmax$ to $\nico{\xminall}$ which lies inside $[\xmaxmoins,\xmaxplus]$ and then its associated effort is equal to $f(\xmax) - f(\xmin)$, which means that the dynamics of $\xmin$ is lower than or equal to this same value. Additionally, we can show that every optimal path  connects $\xmin$ to $\xmax$ and thus the dynamics of $\xmin$ is greater than or equal to $f(\xmax) - f(\xmin)$. It is then easy to conclude that the dynamics of $\xmin$ is equal to $f(\xmax) - f(\xmin)$, and then by \nico{uniqueness} of the critical values, $\xmin$ is paired with $\xmax$ by dynamics.}
    \label{fig:sketch1D-per-dyn}
\end{figure}

\begin{table*}[!htbp]
\centering
\caption{Sketches of the 1D/$n$-D proofs that pairing by persistence implies pairing by dynamics.}
\begin{tabular}{l|c|l}
\toprule
\multicolumn{3}{c}{Hypotheses:}\\
\toprule
\toprule
$f$ is a \Morse function & & $f$ is a Morse function\\
\midrule
\multicolumn{3}{c}{$f$ has unique critical values}\\
\midrule
\multicolumn{3}{c}{$\XMAXSAD$ is a local maximum/$1$-saddle of $f$}\\
\midrule
\multicolumn{3}{c}{$\XMAXSAD$ and $\xmin$ are paired by \textbf{persistence}}\\
\midrule
$\xmax > \xmin$ & \hspace{1em} & $\xmin \neq \xsad$\\
\toprule
\toprule
\multicolumn{3}{c}{Notations:}\\
\toprule
\toprule
$[\xmaxmoins,\xmaxplus] = \connectedcomponent([f \leq f(\xmax)],\xmax)$ & & $\CSAD = \CC([f \leq f(\xsad)],\xsad)$\\
\midrule
$\nico{\xminall} = \rep([\xmaxmoins,\xmaxplus],f)$ & & $\{\CI\}_{i \in I} = \CoCo([f < f(\xsad)])$\\
\midrule
& & $\{\CSADI\}_{i \in \ISAD} = \left\{ \CI \ |\ \xsad \in \CLO(\CI) \right\}$\\
\midrule
& & $\imin \in \ISAD$ s.t. $\xmin$ represents $\CSADIMIN$\\
\bottomrule
\bottomrule
\multicolumn{3}{c}{Step 1:}\\
\toprule
\toprule
$\gamma := [\xmin,\nico{\xminall}]$ & & $\CARD(\ISAD) > 1$\\
with $f(\nico{\xminall}) < f(\xmin)$ & & $\Rightarrow$ $\exists \ \ILOWER \in \ISAD$, $\exists \ \XLOWER \in \CSADLOWER$,\\
& & \hspace{0.4cm} s.t. $f(\XLOWER) < f(\xmin)$\\
\midrule
\multicolumn{3}{c}{$\xmin$ is matchable}\\
\bottomrule
\bottomrule
\multicolumn{3}{c}{Step 2:}\\
\toprule
\toprule
$\gamma$ is a descending path & & $\exists\ \gamma_1$ from $\xmin$ to $\xsad$ in $\CSADIMIN$\\
& & $\forall\ i \in \ISAD \setminus\{\imin\}, \exists\ \gamma_2$ from $\xsad$ to $\XLOWER$\\
& & $\Rightarrow \gamma := \gamma_1 <>\gamma_2$ is a descending path\\
\midrule
\multicolumn{3}{c}{the dynamics of $\gamma$ is equal to $f(\xmax) - f(\xmin)$}\\
\multicolumn{3}{c}{$\Rightarrow \DYN(\xmin) \leq f(\XMAXSAD) - f(\xmin)$}\\
\bottomrule
\bottomrule
\multicolumn{3}{c}{Step 3:}\\
\toprule
\toprule
If $\xmin$ is paired by dynamics with $\xstar$ & & $\DYN(\xmin) < f(\xsad) - f(\xmin)$ $\HYPO$\\
Then $\xstar > \xmin$ & & $ \Rightarrow \exists$ a descending $\gamma$ from $\xmin$ in $\CSADIMIN$\\
$\XLOWER := \inf\{x > \xmin \; ; \; f(x) < f(\xmin)\}$ & & $\Rightarrow \xmin$ does not represent $\CSADIMIN$\\
$\XLOWER > \xmax$ & & $\Rightarrow \HYPO$ is false\\
$\gamma$ optimal path $\Rightarrow$ $\{\xmin, \xmax, \XLOWER\} \in \gamma$ & & \\
\midrule
\multicolumn{3}{c}{$\DYN(\xmin) \geq f(\XMAXSAD) - f(\xmin)$}\\
\bottomrule
\bottomrule
\multicolumn{3}{c}{Step 4:}\\
\toprule
\toprule
\multicolumn{3}{c}{$\DYN(\xmin) = f(\XMAXSAD) - f(\xmin)$}\\
\midrule
\multicolumn{3}{c}{$\XMAXSAD$ and $\xmin$ are paired by dynamics}\\
\bottomrule
\end{tabular}
\label{table.sketch.per.dyn}
\end{table*}

We assume as usual that $f$ is a Morse function (see Table~\ref{table.sketch.per.dyn}), that its critical values are unique. Let us prove that when some maximum of $f$ in the 1D case (or some $1$-saddle of $f$ in the $n$-D case) is paired by persistence to some minimum of this same function $f$, then this minimum is paired with this maximum (resp. this $1$-saddle) by dynamics.

\medskip

\textbf{\underline{1D proof:}} Let us start with the 1D case (see Figure~\ref{fig:sketch1D-per-dyn}). By considering that some maximum $\xmax$ is paired with some minimum $\xmin$ by persistence (with $\xmin < \xmax$), we obtain at the same time several properties (by definition of the pairing by persistence): 

\begin{itemize}

    \item we can draw the threshold set $[f \leq f(\xmax)]$ at level $f(\xmax)$,
    
    \item we know that it draws a connected component $$\connectedcomponent([f \leq f(\xmax)],\xmax)$$ containing $\xmax$ that we can define as $[\xmaxmoins,\xmaxplus]$ with $\xmaxmoins < \xmax < \xmaxplus$, 
    
    \item we know then that $\xmin$ is the representative of $[\xmaxmoins,\xmax]$ and we can define some $\nico{\xminall}$ as being the representative of $[\xmax,\xmaxplus]$, with $f(\xmin) > f(\nico{\xminall})$.
    
\end{itemize}

Now let us prove that $\xmin$ is paired by dynamics to $\xmax$ in four steps. First, we know that there exists some path $\gamma : [0,1] \rightarrow [\xmin,\nico{\xminall}] : \lambda \rightarrow (1-\lambda) \xmin + \lambda \nico{\xminall}]$ joining $\xmin$ to $\nico{\xminall}$ with $f(\nico{\xminall}) < f(\xmin)$, then $\xmin$ is matchable.

\medskip

Then, the second step is straightforward: since $\gamma$ reaches some $\nico{\xminall}$ with an altitude lower than the one of $\xmin$, it is a descending path. Furthermore, the effort associated to $\gamma$ is equal to $f(\xmax) - f(\xmin)$, since we have to reach $(\xmax,f(\xmax))$ when we start from $(\xmin,f(\xmin))$ to be able to go down to $$(\nico{\xminall},f(\nico{\xminall})).$$ Then the optimal effort associated to $\xmin$, that is the dynamics of $\xmin$, is lower than or equal to $f(\xmax) - f(\xmin)$.

\medskip

Now, for the third step, we assume that $\xmin$ is paired with some $\xstar < \xmin$, which is clearly impossible: otherwise dynamics of $\xmin$ would be greater than $f(\xmax) - f(\xmin)$ (we would need to go outside the connected component $[\xmaxmoins,\xmaxplus]$ to reach some altitude lower than $f(\xmin)$). Then $\xmin$ is paired with some maximum $\xstar$ greater than $\xmin$. Now, we define $\XLOWER$ as the \textquote{first} abscissa of altitude lower than $f(\xmin)$ on the right side of $\xmin$; obviously this abscissa is greater than $\xmax$ since $\xmin$ is the representative of the basin $[\xmaxmoins,\xmax]$. Since any optimal descending path starting from $\xmin$  goes through the abscissas $\xmin$, $\xmax$ and then $\XLOWER$, its associated effort is greater than or equal to $f(\xmax) - f(\xmin)$.

\medskip

The fourth step combines the previous properties and leads to the conclusion that the dynamics of $\xmin$ is equal to $f(\xmax) - f(\xmin)$, which means that the maxima associated to $\xmin$ by dynamics is $\xmax$ (by \nico{uniqueness} of the critical values).

\medskip

\textbf{\underline{$n$-D proof:}} The main steps of the $n$-D proof are very similar to the 1D case. However, the notations are very different, due to the fact that the number of path from one point to another in $\Reals^n$ is infinite (and there is no \textquote{left} nor \textquote{right}). Starting from the $1$-saddle $\xsad$ paired by persistence to $\xmin$, we have to use the following notations:

\begin{itemize}
    
    \item we define the closed component $\CSAD = \connectedcomponent([f \leq f(\xsad)],\xsad)$,
    
    \item we define also the open components $\{\CIMIN\}_i$ of $[f < f(\xsad)]$, whose subset $\{\CSADI\}_i$ corresponds to these components whose closure contains $\xsad$,
    
    \item we call $\imin$ the index of the component $\CSADIMIN$ that $\xmin$ represents.
    
\end{itemize}

\medskip

The first step consists of recalling that the number of components of $\CSADI$ is equal to two, then greater than one, and thus there exists some index $\ILOWER$ and some abscissa $\XLOWER \in \CSADLOWER$ such that $f(\XLOWER) < f(\xsad)$ (since pairing by persistence associates $\xsad$ to the local minimum of the highest altitude). Thus, $\xmin$ is matchable.

\medskip

As a second step, we construct a path $\gamma_1$ from $\xmin$ to $\xsad$ in $\CSADIMIN$ and another path $\gamma_2$ from $\xsad$ to $\XLOWER$ in the component $\CSADI$ containing it, from which we deduce a descending path $\gamma := \gamma_1 <> \gamma_2$ associated to $\xmin$. Thus, the effort associated to $\gamma$ is lower than or equal to $f(\xsad) - f(\xmin)$ (since this path has not yet been shown to be optimal).

\medskip

The third step uses a proof by contradiction. We assume that the dynamics of $\xmin$ is lower than $f(\xsad) - f(\xmin)$; we call this hypothesis $\HYP$. Then, $\HYP$ implies that there exists a descending path inside the component $\CSADIMIN$, which implies that $\xmin$ does not represent $\CSADIMIN$, which is impossible (it contradicts the hypotheses). Then, the dynamics of $\xmin$ is greater than or equal to $f(\xsad) - f(\xmin)$.

\medskip 

As for the 1D case, the fourth steps concludes: since the dynamics of $\xmin$ is equal to $f(\xsad) - f(\xmin)$ thanks to the combination of the previous steps, the only possible local maximum paired by dynamics to $\xmin$ is $\xsad$.

\section{Pairings by dynamics and by persistence are equivalent in 1D}
\label{sec.equivalence1D}

In this section, we prove that under some constraints, pairings by dynamics and by persistence are equivalent in the 1D case.

\begin{figure}
\centering
\includegraphics[width=0.8\linewidth]{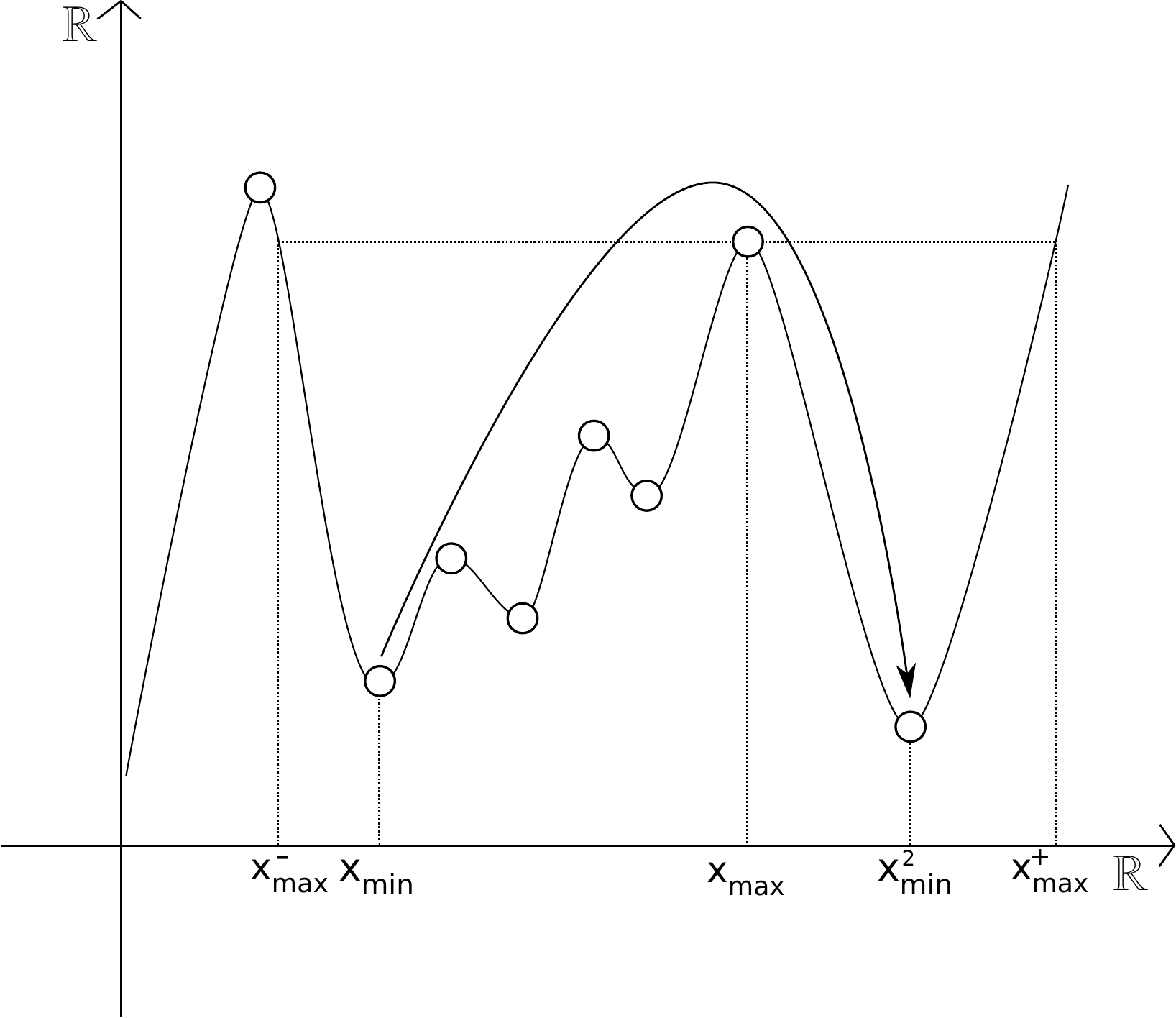}
\caption{A \Morse function where the local extrema $\xmin$ and $\xmax$ are paired by dynamics.}
\label{fig.proofDYNPER}
\end{figure}

\begin{Proposition}
Let $f : \Reals \rightarrow \Reals$ be a \Morse function with unique critical values. Now, let us assume that a local minimum $\xmin \in \Reals$ of $f$ is paired with a local maximum $\xmax$ of $f$ by dynamics. We assume \nico{without loss of generality} that $\xmin < \xmax$ (the reasoning is the same for the opposite assumption). Also, we denote by $(\xmax^-,\xmax^+) \in \realscomplete^2$ the two values verifying: $$[\xmax^-,\xmax^+] = \cloRB (\connectedcomponent([f \leq f(\xmax)],\xmax)).$$
Then the following properties are true:
\begin{itemize}
\setlength\itemsep{1em}
\item[(P1)] $\xmin = \rep([\xmax^-,\xmax],f)$,
\item[(P2)] When $\xmax^+$ is finite, $\xmindeux := \rep([\xmax,\xmax^+],f)$ \nico{satisfies} $f(\xmindeux) < f(\xmin)$,
\item[(P3)] $\xmax$ and $\xmin$ are paired by persistence.
\end{itemize}
\label{proposition.DYNPER}
\end{Proposition}

\Proof Figure~\ref{fig.proofDYNPER} depicts an example of \Morse function where $\xmin$ and $\xmax$ are paired by dynamics.

\medskip

\begin{figure}
\centering
\includegraphics[width=0.8\linewidth]{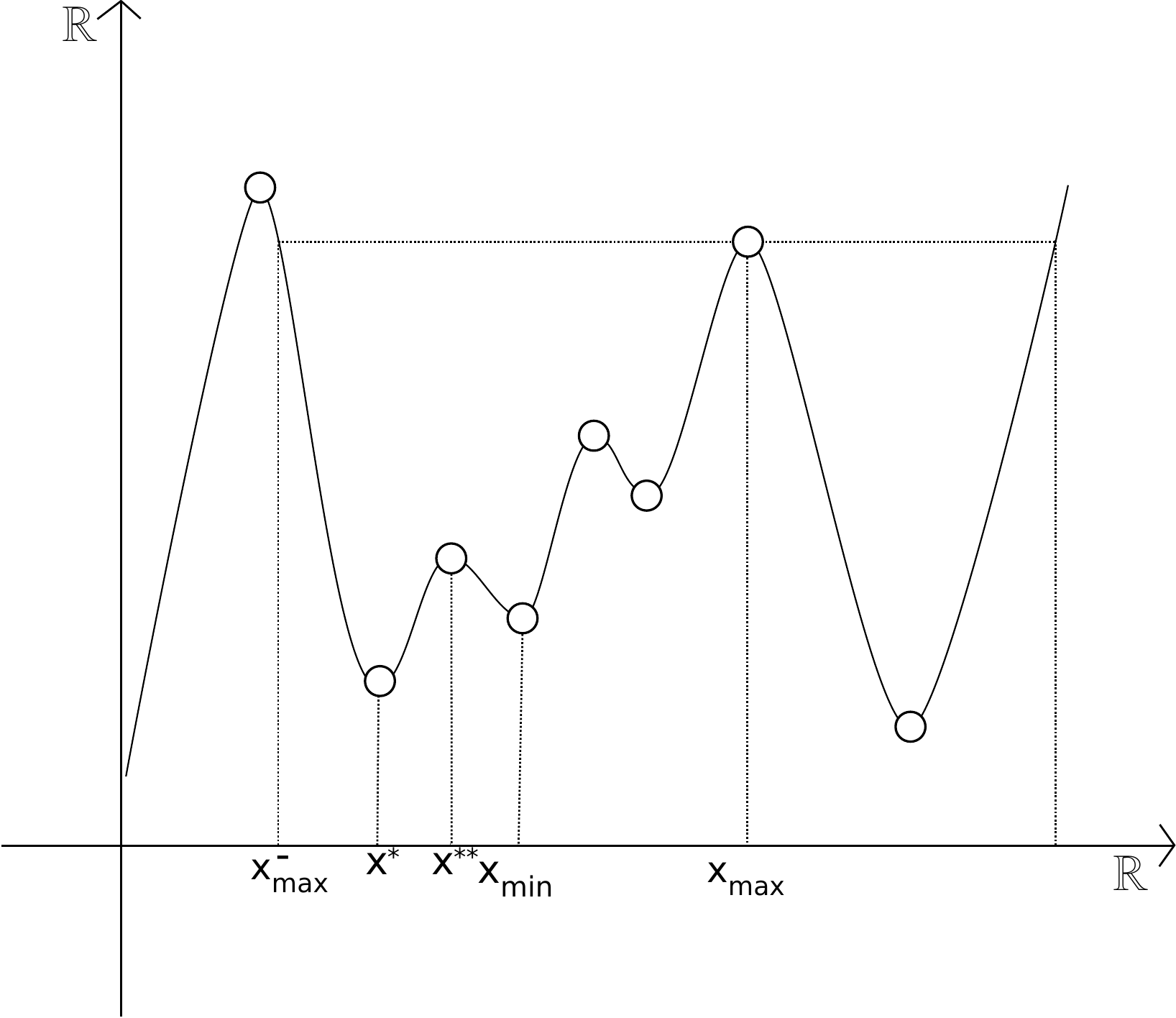}
\caption{Proof of $(P1)$.}
\label{fig:absurde1}
\end{figure}

Let us prove $(P1)$; we proceed by \textit{reductio ad absurdum}. When $\xmin$ is not the lowest local minimum of $f$ on the interval $[\xmax^-,\xmax]$, then there exists another local minimum $x^* \in [\xmax^-,\xmax]$ of $f$ (see Figure~\ref{fig:absurde1}) which \nico{satisfies} $f(x^*) < f(\xmin)$ ($x^*$
 and $\xmin$ being distinct local extrema of $f$, their images by $f$ are not equal). Then, because the path joining $\xmin$ and $x^*$ belongs to $C$ (defined in Subsection~\ref{ssec.dyn}), we have: $$\DYN(\xmin) \leq \max\{f(x) - f(\xmin) \; ; \; x \in \iv(x^*,\xmin)\}.$$
Let us call $x^{**} := \arg \max_{x \in [\iv(\xmin,x^*)]} f(x)$, we can deduce that $f(x^{**}) < f(\xmax)$ since $x^{**} \in \iv(x^*,\xmin) \subseteq ]\xmax^-, \xmax[$. In this way, $$\DYN(\xmin) \leq f(x^{**}) - f(\xmin),$$ which is lower than $f(\xmax) - f(\xmin)$; this is a contradiction since $\xmin$ and $\xmax$ are paired by dynamics. $(P1)$ is then proved.

\medskip

\begin{figure}
\centering
\includegraphics[width=0.8\linewidth]{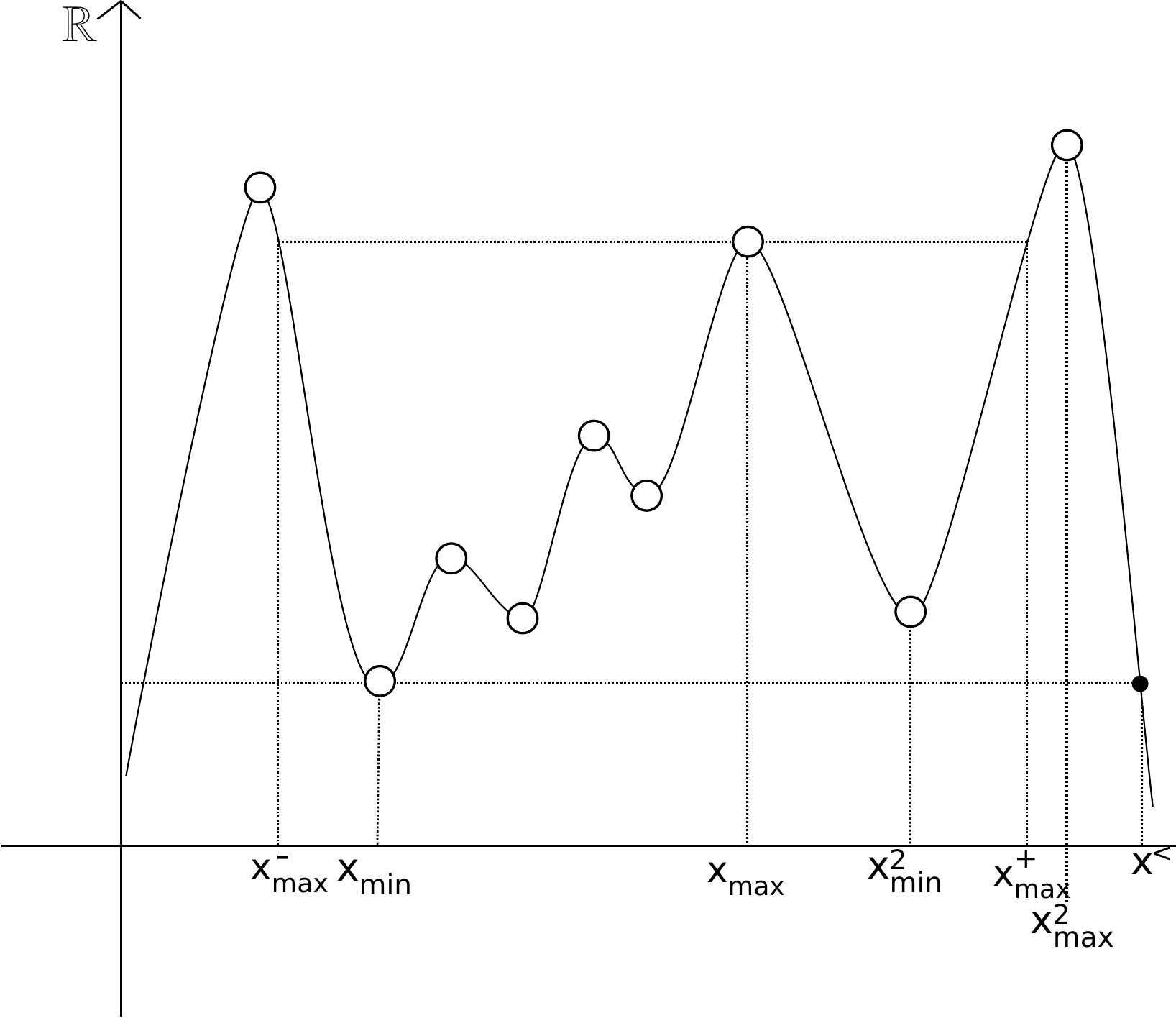}
\caption{Proof of $(P2)$ in the case where $\xmax^+$ is finite.}
\label{fig:absurde2}
\end{figure}

Now let us prove $(P2)$. Let us assume that $\xmax^+$ is finite and let $\xmindeux$ be the representative of $[\xmax,\xmax^+]$ relatively to $f$. Let us assume that $f(\xmindeux) > f(\xmin)$. Note that we cannot have equality of $f(\xmindeux)$ and $f(\xmin)$, since $\xmin$ and $\xmindeux$ are both local extrema of $f$. Then we obtain Figure~\ref{fig:absurde2}. Since with $x \in [\xmax,\xmax^+]$, we have $f(x) \geq f(\xmindeux) > f(\xmin)$, and because $\xmin$ is paired with $\xmax$ by dynamics with $\xmin < \xmax$, then there exists a value $x$ on the right of $\xmax$ where $f(x)$ is lower than $f(\xmin)$. In other words, there exists: $$x^< := \inf\{ x \in [\xmax,+\infty[ \; ; \; f(x) < f(\xmin)\}$$ such that for some arbitrarily small value $\varepsilon > 0$, $f(x^< + \varepsilon) < f(\xmin)$. Since $x^< > \xmax^+$, any path $\gamma$ joining $\xmin$ to $x^<$ goes through a local maximum $\xmaxdeux$ defined by
$$\xmaxdeux := \arg \max_{x \in [\xmax^+,x^<]} f(x)$$
which \nico{satisfies} $f(\xmaxdeux) > f(\xmax^+)$. Then the dynamics of $\xmin$ is greater than or equal to $f(\xmaxdeux) - f(\xmin)$ which is greater than $f(\xmax) - f(\xmin)$. We obtain a contradiction. Then we have $f(\xmindeux) < f(\xmin)$. The proof of $(P2)$ is done.

\medskip

Thanks to $(P1)$ and $(P2)$, we obtain directly $(P3)$ by applying Algorithm~\ref{algo.pairingbypersistence}. \qed

\begin{Proposition}
Let $f : \Reals \rightarrow \Reals$ be a \Morse function with unique critical values. Now, let us assume that a local minimum $\xmin \in \Reals$ of $f$ is paired with a local maximum $\xmax$ of $f$ by persistence. We assume \nico{without loss of generality} that $\xmin < \xmax$ (the reasoning is the same for the opposite assumption). Then, $\xmax$ and $\xmin$ are paired by dynamics.
\label{proposition.PERDYN}
\end{Proposition}

\begin{figure}
\centering
\includegraphics[width=0.8\linewidth]{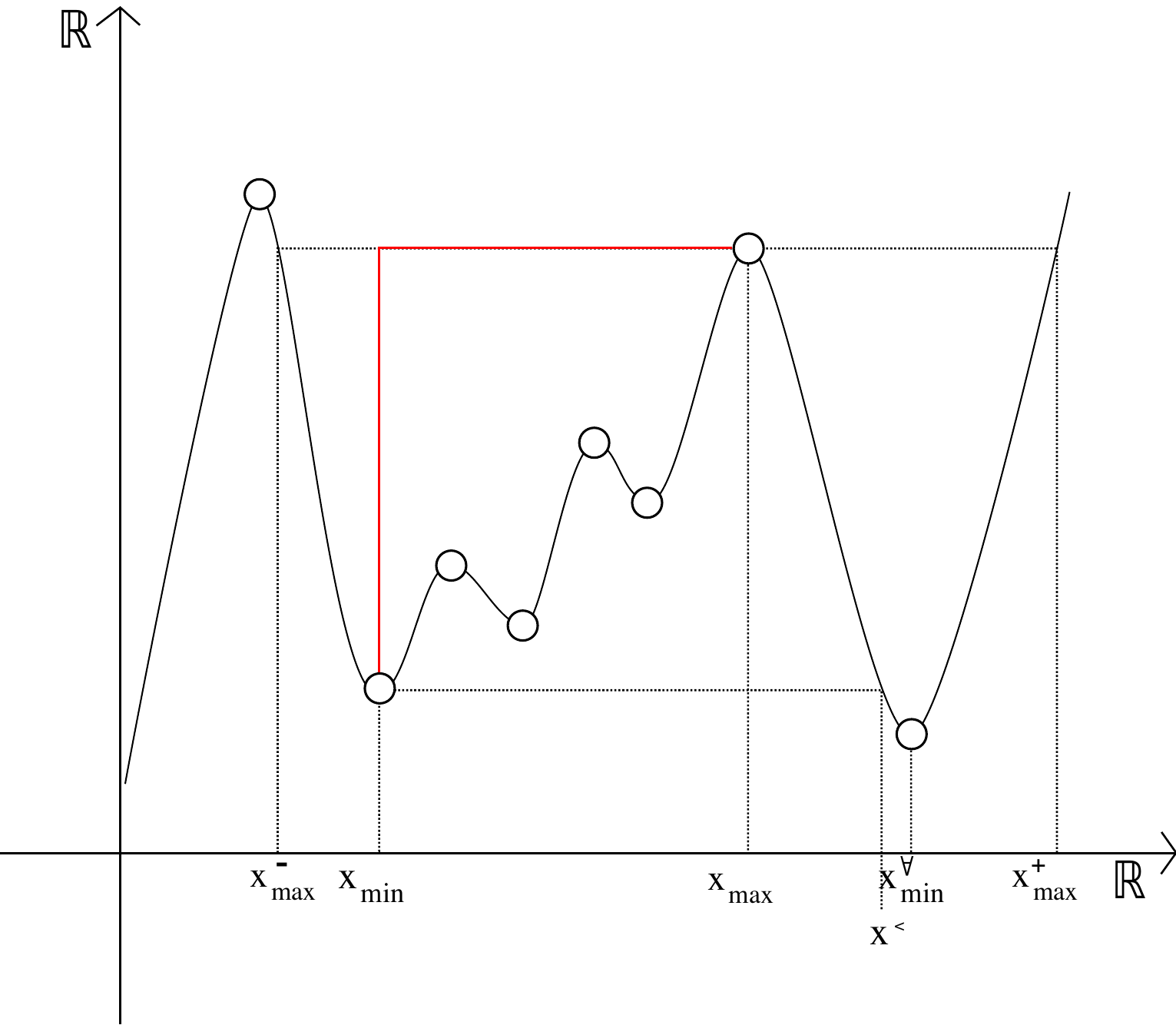}
\caption{A \Morse function $f : \Reals \rightarrow \Reals$ where the local extrema $\xmin$ and $\xmax$ are paired by persistence relatively to $f$.}
\label{fig.proofPERDYN}
\end{figure}

\Proof We denote by $(\xmax^-,\xmax^+) \in \realscomplete^2$ the two values verifying: $$[\xmax^-,\xmax^+] = \cloRB(\connectedcomponent([f \leq f(\xmax)],\xmax)).$$

Since $\xmin$ is paired by persistence to $\xmax$ with $\xmin < \xmax$ (see Figure~\ref{fig.proofPERDYN}), then:
$$\xmin = \rep([\xmax^-,\xmax],f) \in \Reals,$$
and, by Algorithm~\ref{algo.pairingbypersistence}, we know that $\xmax^- > - \infty$ (then $\xmax^-$ is finite).

\medskip

When $\xmax^+ < +\infty$ (Case $1$), the representative $\nico{\xminall}$ of $[\xmax,\xmax^+]$ relatively to $f$ is exists in $]\xmax,\xmax^+[$ and is unique, and its image by $f$ is lower than $f(\xmin)$. When $\xmax^+ = +\infty$ (Case $2$), $\lim_{x \rightarrow +\infty} f(x) = -\infty$, and then there exists one more time an abscissa $\nico{\xminall} \in \Reals$ whose image by $f$ is lower than $f(\xmin)$. So, in both cases, there exists a (finite) value $\nico{\xminall} \in ]\xmax,\xmax^+[$ verifying $f(\nico{\xminall}) < f(\xmin)$. This way, we know that $\xmin$ is paired with some abscissa in $\Reals$ by dynamics.

\medskip

In Case $1$, we know that the path defined as:
$$\gamma : \lambda \in [0,1] \rightarrow \gamma(\lambda) := (1-\lambda) \xmin + \lambda \nico{\xminall}$$
belongs to the set of paths $C$ defining the dynamics of $\xmin$ (see Subsection~\ref{ssec.dyn}). Then, $$\DYN(\xmin) \leq \max\{f(x) - f(\xmin) \; ; \; x \in \gamma([0,1])\},$$
which is lower than or equal to $f(\xmax) - f(\xmin)$ since $f$ is maximal at $\xmax$ on $[\xmax^-,\xmax^+]$. Then we have the following property: $$\DYN(\xmin) \leq f(\xmax) - f(\xmin).$$ In Case $2$, since $f(x)$ is lower than $f(\xmax)$ for $x \in ]\xmax,+\infty[$, then one more time we get $\DYN(\xmin) \leq f(\xmax) - f(\xmin)$. Let us call this property $(P)$.

\medskip

\begin{figure}
\centering
\includegraphics[width=0.8\linewidth]{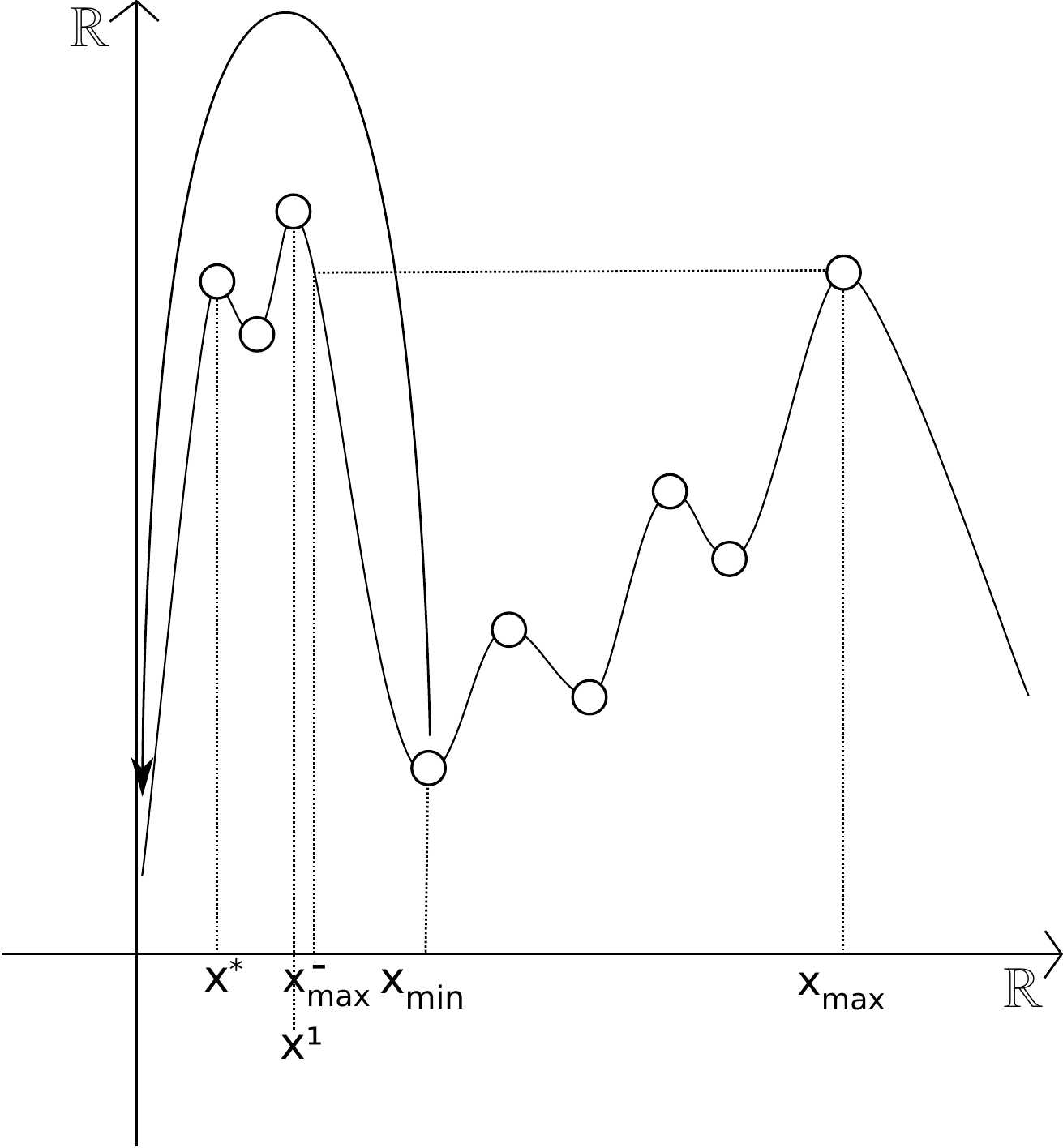}
\caption{The proof that it is impossible to obtain a local maximum $x^* < \xmin$ paired with $\xmin$ by dynamics when $\xmin$ is paired with $\xmax > \xmin$ by persistence.}
\label{fig:absurde3}
\end{figure}

Even if we know that there exists some local maximum of $f$ which is paired with $\xmin$ by dynamics, we do not know whether the abscissa of this local maximum is lower than or greater than $\xmin$. Then, let us assume that there exists a local maximum $x^* < \xmin$ (lower case) which is associated to $\xmin$ by dynamics. We denote this property $(H)$ and we depict it in Figure~\ref{fig:absurde3}. Since $f(x)$ is greater than or equal to $f(\xmin)$ for $x \in [\xmax^-,\xmin]$, $(H)$ implies that $x^* < \xmax^-$. Then, we can observe that the local maximum $x^1$ of $f$ of maximal abscissa in $[x^*,\xmax^-]$ \nico{satisfies} $f(x^1) > f(\xmax)$, which implies that $\DYN(\xmin) \geq f(x^1) - f(\xmin) > f(\xmax) - f(\xmin)$ (since we go through $x^1$ to reach $x^*$), which contradicts $(P)$. $(H)$ is then false. In other words, we are in the upper case: the local maximum paired by dynamics to $\xmin$ belongs to $]\xmin, +\infty[$, let us call this property $(P')$.

\medskip

Now let us define: $$x^< := \inf\{x > \xmin \; ; \; f(x) < f(\xmin)\},$$ (see again Figure~\ref{fig.proofPERDYN}) and let us remark that $x^< > \xmax$ (because $\xmin$ is the representative of $f$ on $[\xmax^-,\xmax]$). Since we know by $(P')$ that a local maximum $x > \xmin$ of $f$ is paired by dynamics with $\xmin$, then the image of every optimal path belonging to $C$ contains $\{x^<\}$, and then contains $[\xmin,x^<]$. Indeed, an optimal path in $C$ whose image would not contain $\{x^<\}$ would then contain an abscissa $x < \xmax^-$ and then we would obtain $\DYN(\xmin) > f(\xmax) - f(\xmin)$, which would contradict $(P)$.

\medskip

Now, the maximal value of $f$ on $[\xmin,x^<]$ is equal to $f(\xmax)$, then $\DYN(\xmin) = f(\xmax) - f(\xmin)$. The only local maximum of $f$ whose value is $f(\xmax)$ is $\xmax$, then $\xmax$ is paired with $\xmin$ by dynamics relatively to $f$. \qed

\begin{Th}
Let $f : \Reals \rightarrow \Reals$ be a \Morse function with a finite number of local extrema and unique critical values. A local minimum $\xmin \in \Reals$ of $f$ is paired by dynamics to a local maximum $\xmax \in \Reals$ of $f$ iff $\xmax$ is paired by persistence to $\xmin$. In other words, pairings by dynamics and by persistence lead to the same result. Furthermore, we obtain $\per(\xmax) = \DYN(\xmin)$.
\label{th.equivalent}
\end{Th}

\Proof This theorem results from Propositions~\ref{proposition.DYNPER} and~\ref{proposition.PERDYN}. \qed

\medskip

Note that pairing by persistence has been proved to be \emph{symmetric} in~\cite{cohen2009extending} for Morse functions defined on manifolds: the pairing is the same for a Morse function and its negative.

\section{The $n$-D equivalence}
\label{sec.equivalence}

Let us make two important remarks that will help us in the sequel.

\begin{figure}
\centering
\includegraphics[width=0.7\linewidth]{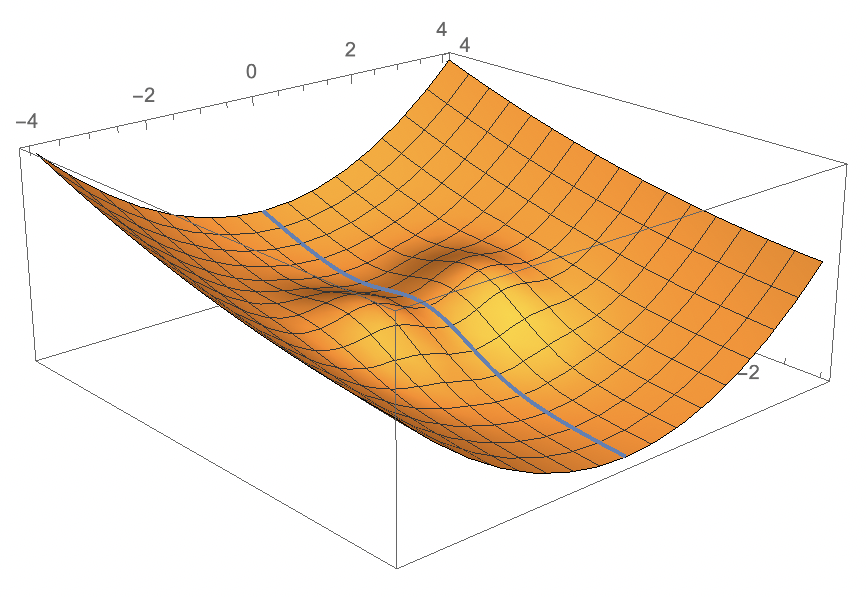}
\caption{Every optimal descending path goes through a $1$-saddle. Observe the path in blue coming from the left side and decreasing when following the topographical view of the Morse function $f$. The effort of this path to reach the minimum of $f$ is minimal thanks to the fact that it goes through the saddle point at the middle of the image.}
\label{fig.optsad}
\end{figure}

\begin{Lem}
\label{lemma.saddle}
Let $f : \Reals^n \rightarrow \Reals$ be a Morse function and let $\xmin$ be a local minimum of $f$. Then for any optimal path $\PIOPT$ in $\DXMIN$, there exists some $\lstar \in ]0,1[$ such that it is a maximum of $f \circ \PIOPT$ and at the same time $\PIOPT(\lstar)$ is the abscissa of a $1$-saddle point of $f$.
\end{Lem}

\ProofNico: This proof is depicted in Figure~\ref{fig.optsad}. Let us proceed by counterposition, and let us prove that when a path $\Pi$ in $\DXMIN$ does not go through a $1$-saddle of $f$, it cannot be optimal.

\medskip

Let $\Pi$ be a path in $\DXMIN$. Let us define $\lstar \in [0,1]$ as one of the positions where the mapping $f \circ \Pi$ is maximal:
$$\lstar \in {\arg \max}_{\ell \in [0,1]} f( \Pi (\ell)),$$
and $\xstar = \Pi(\lstar)$. Let us prove that we can find another path $\Pi'$ in $\DXMIN$ whose effort is lower than the one of $\Pi$.

\medskip

\begin{figure}
\centering
\includegraphics[width=0.45\linewidth]{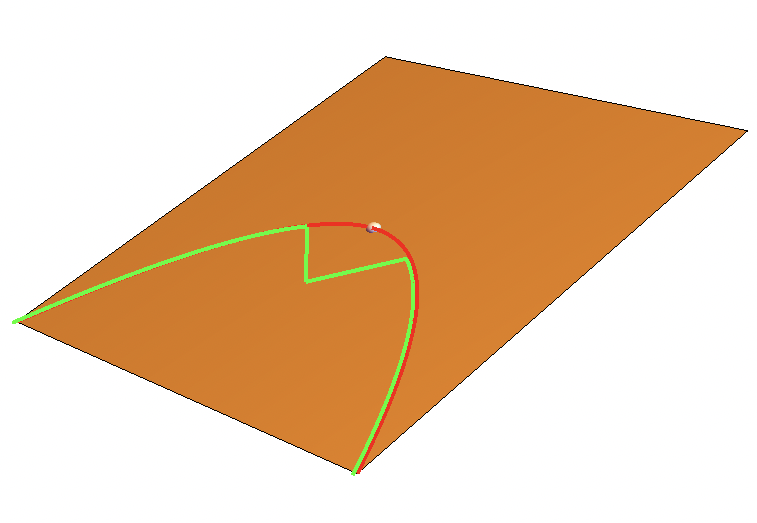}~\includegraphics[width=0.3\linewidth]{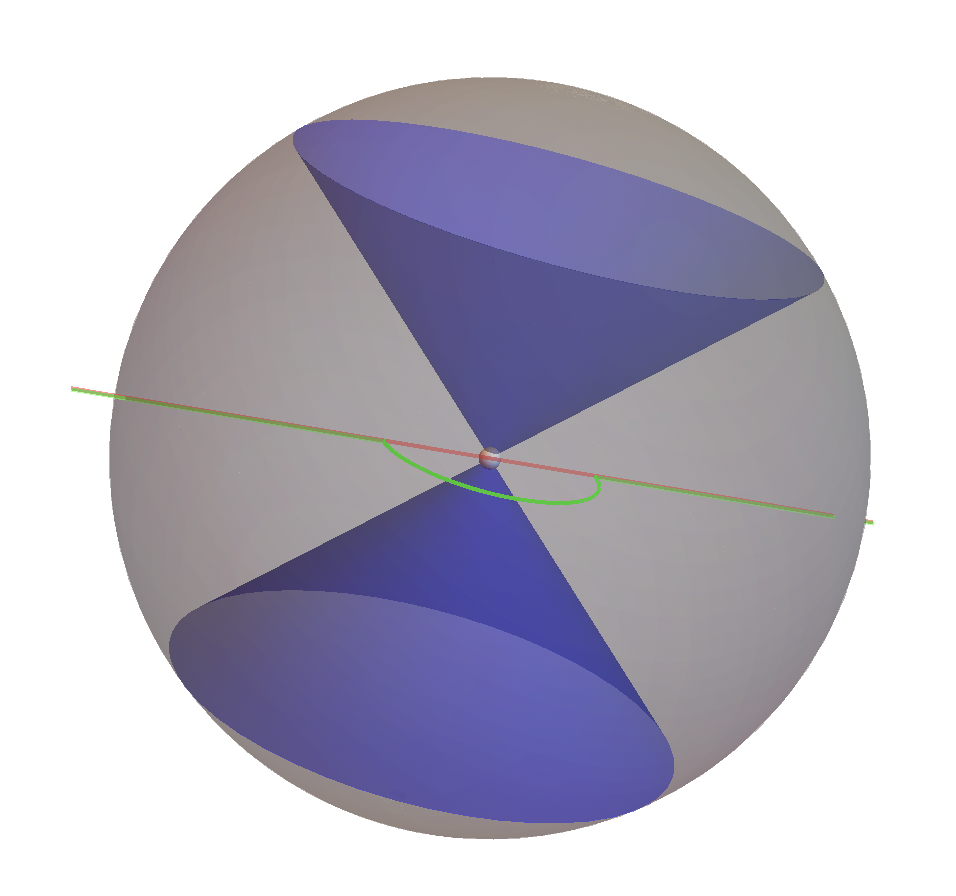}
\caption{How to compute descending paths of lower efforts. The initial path going through $\xstar$ (the little gray ball) is in red, the new path of lower effort is in green (the non-zero gradient case is on the left side, the zero-gradient case is on the right side).}
\label{fig.lower}
\end{figure}
    
At $\xstar$, $f$ can satisfy three possibilities:

\begin{itemize}
    
    \item When we have $\nabla f (\xstar) \neq 0$ (see the left side of Figure~\ref{fig.lower}), then locally $f$ is a plane of slope $\|\nabla f (\xstar)\|$, and then we can easily find some path $\Pi'$ in $\DXMIN$ with a lower effort than $\effort(\Pi)$. More precisely, let us fix some arbitrary small value $\varepsilon > 0$ and draw the closed topological ball $\VOISINAGEXSTAR$, we can define three points: 
\begin{align*}
\lmin & = \min\{\ell \ |\ \Pi(\ell) \in \VOISINAGEXSTAR\},\\
\lmax & = \max\{\ell \ |\ \Pi(\ell) \in \VOISINAGEXSTAR\},\\
x_B & = \xstar - \varepsilon . \frac{\nabla f (\xstar)}{\|\nabla f (\xstar)\|}.
\end{align*}

    Thanks to these points, we can define a new path $\Pi'$:

\scalebox{0.85}{\parbox{\linewidth}{%
$$\Pi|_{[0,\lmin]} <> [\Pi(\lmin),x_B] <> [x_B,\Pi(\lmax)] <> \Pi|_{[\lmax,1]}.$$
}}

    By doing this procedure at every point in $[0,1]$ where $f \circ \Pi$ reaches its maximal value, we obtain a new path whose effort is lower than the one of $\Pi$.
    
    \item When we have $\nabla f (\xstar) = 0$, then we are at a critical point of $f$. It cannot be a $0$-saddle, that is, a local minimum, due to the existence of the descending path going through $\xstar$. It cannot be a $1$-saddle neither (by hypothesis). It is then a $k$-saddle point with $k \in [2,n]$ (see the right side of Figure~\ref{fig.lower}). Using Lemma~\ref{lemma.morse}, $f$ is locally equal to a second order polynomial function (up to a change of coordinates $\varphi$ s.t. $\varphi(\xstar) = \veczero$): $$ \ f \circ \varphi^{-1} (\vecx) = f(\xstar) - x_1^2 - x_2^2 - \dots - x_k^2 + x_{k+1}^2 + \dots + x_n^2.$$
    Now, let us define for some arbitrary small value $\varepsilon > 0$:
\begin{align*}
\lmin & = \min\{\ell \ |\ \Pi(\ell) \in \VOISINAGEZERO\},\\
\lmax & = \max\{\ell \ |\ \Pi(\ell) \in \VOISINAGEZERO\},\\
\end{align*}
and 

\scalebox{0.85}{\parbox{\linewidth}{%
$$\BFRAK = \left\{\vecx \ \Big| \ \sum_{i \in [1,k]} x_i^2 \leq \varepsilon^2 \ \textbf{ and } \forall j \in [k+1,n], x_j = 0\right\} \setminus \{\veczero\}.$$
}}

    This last set is connected since it is equal to a $k$-manifold (with $k \geq 2$) minus a point. Let us assume \nico{without loss of generality} that $\Pi(\lmin)$ and $\Pi(\lmax)$ belong to $\BFRAK$ (otherwise we can consider their orthogonal projections on the hyperplane of lower dimension containing $\BFRAK$ but the reasoning is the same). Thus, there exists some path $\Pi_\BFRAK$ joining $\Pi(\lmin)$ to $\Pi(\lmax)$ in $\BFRAK$, from which we can deduce the path $\Pi' = \Pi|_{[0,\lmin]} <> \Pi_\BFRAK <> \Pi|_{[\lmax,1]}$ whose effort is lower than the one of $\Pi$ since its image is inside $[f < f(\xstar)]$.
    
\end{itemize}

Since we have seen that, in any possible case, $\Pi$ is not optimal, it concludes the proof.\qed

\begin{figure}
\centering
\includegraphics[width=0.4\linewidth]{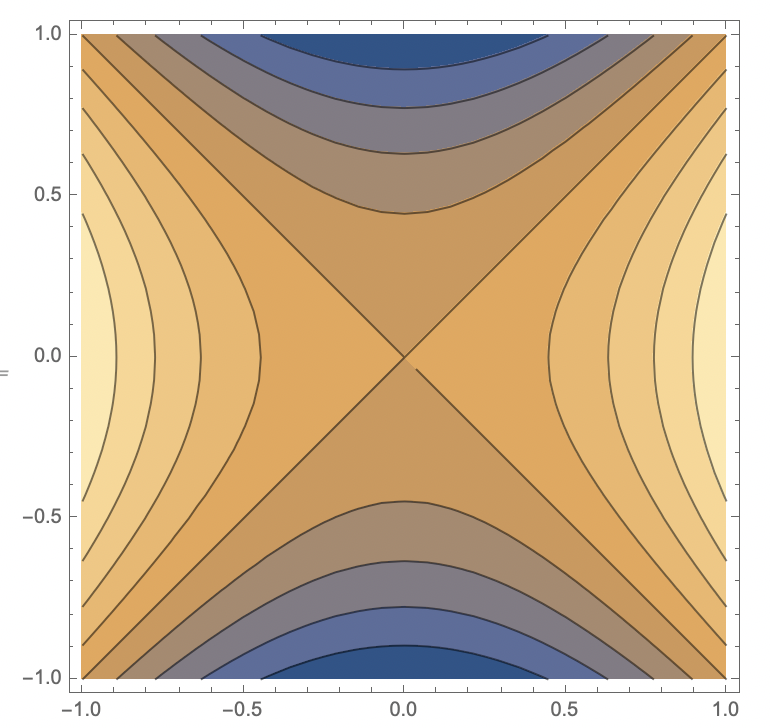}
\caption{A $1$-saddle point leads to two open connected components. At a $1$-saddle point whose abscissa is $\xsad$ (at the center of the image), the component $[f \leq f(\xsad)]$ is locally the merge of the closure of two connected components (in orange) of $[f < f(\xsad)]$ when $f$ is a Morse function.}
\label{fig.saddle}
\end{figure}

\begin{Proposition}
\label{proposition.2}
Let $f$ be a Morse function from $\Reals^n$ to $\Reals$ with $n \geq 1$. When $\xstar$ is a critical point of index $1$, then there exists $\varepsilon > 0$ such that:
$$\CARD\left(\CC(B(\xstar,\varepsilon) \cap [f < f(\xstar)])\right) = 2,$$
where $\CARD$ is the \emph{cardinality operator}.
\end{Proposition}

\ProofNico: The case $n = 1$ is obvious, let us then treat the case $n \geq 2$ (see Figure~\ref{fig.saddle}). Thanks to Lemma~\ref{lemma.morse} and thanks to the fact that $\xsad$ is the abscissa of a $1$-saddle, we can say that (up to a change of coordinates and in a small neighborhood around $\xsad$) for any $\vecx$: 

$$f(x) = f(\xsad) + \vecx^T . \MATRICEHESSENNESADDLE . \vecx,$$

where $\IDENTITYMATRIXNMOINSUN$ is the identity matrix of dimension $(n-1) \times (n-1)$. In other words, around $\xsad$, we obtain that:

$$[f < f(\xsad)] = \left\{\vecx \ \Big| \ - x_1^2 + \sum_{i = 2}^n x_i^2 < 0 \right\} = \CPLUS \cup \CMINUS,$$

with:

\begin{align*}
\CPLUS & = \left\{\vecx \ \Big| \ x_1 > \sqrt{\sum_{i = 2}^n x_i^2} \right\},\\
\text{and}\\
\CMINUS & = \left\{\vecx \ \Big| \ x_1 < - \sqrt{\sum_{i = 2}^n x_i^2} \right\},
\end{align*}

where $\CPLUS$ and $\CMINUS$ are two open connected components of $\Reals^n$. Indeed, for any pair $(M,M')$ of $\CPLUS$, we have $\XUNM > \sqrt{\sum_{i = 2}^n \XIM^2}$ and $\XUNMPRIME > \sqrt{\sum_{i = 2}^n \XIMPRIME^2}$, from which we define $N = (\XUNM,0,\dots,0)^T \in \CPLUS$ and $N' = (\XUNMPRIME,0,\dots,0)^T \in \CPLUS$ from which we deduce the path $[M,N] <> [N,N'] <> [N',M']$ joining $M$ to $M'$ in $\CPLUS$. The reasoning with $\CMINUS$ is the same. Since $\CPLUS$ and $\CMINUS$ are two connected (separated) disjoint sets, the proof is done.\qed 

\subsection{Pairing by persistence implies pairing by dynamics in $n$-D}

\begin{Th}
Let $f$ be a Morse function from $\Reals^n$ to $\Reals$. We assume that the $1$-saddle point of $f$ whose abscissa is $\xsad$ is paired by persistence to a local minimum $\xmin$ of $f$. Then, $\xmin$ is paired by dynamics to $\xsad$.
\end{Th}

\ProofNico: Let us assume that $\xsad$ is paired by persistence to $\xmin$, then we have the hypotheses described in Definition~\ref{def.persistence}. Let us denote by $\CMIN$ the connected component in $\{C_i\}_{i \in \ISAD}$ satisfying that $\xmin = \rep(C_{\imin})$.

\medskip

Since $\xsad$ is the abscissa of a $1$-saddle, by Proposition~\ref{proposition.2}, we know that $\CARD(\ISAD) = 2$, then there exists: $\XLOWER = \rep(\CLOWER)$ with $\CLOWER$ the component $C_i$ with $i \in I \setminus \{\imin\}$, then $\xmin$ is matchable. Let us assume that the dynamics of $\xmin$ satisfies:
$$\dynamics(\xmin) < f(\xsad) - f(\xmin). \ \ \ (\HYP)$$
This means that there exists a path $\PIINF$ in $\DXMIN$ such that:
$$\max_{\ell \in [0,1]} f(\PIINF(\ell)) - f(\xmin) < f(\xsad) - f(\xmin),$$
that is, for any $\ell \in [0,1]$, $f(\PIINF(\ell)) < f(\xsad)$, and then by continuity in space of $\PIINF$, the image of $[0,1]$ by $\PIINF$ is in $\CMIN$. Because $\PIINF$ belongs to $\DXMIN$, there exists then some $\XLOWER \in \CMIN$ satisfying $f(\XLOWER) < f(\xmin)$. We obtain a contradiction, $(\HYP)$ is then false. Then, we have $\dynamics(\xmin) \geq f(\xsad) - f(\xmin).$

\medskip

Because for any $i \in \ISAD$, $\xsad$ is an accumulation point of $C_i$ in $\Reals^n$, there exist a path $\Pi_m$ from $\xmin$ to $\xsad$ such that:
\begin{align*}
\forall \ell \in [0,1], & \Pi_m(\ell) \in \CSAD,\\
\forall \ell \in [0,1[, & \Pi_m(\ell) \in \CMIN.
\end{align*}

In the same way, there exists a path $\Pi_M$ from $\XLOWER$ to $\xsad$ such that:
\begin{align*}
\forall \ell \in [0,1], & \Pi_M(\ell) \in \CSAD,\\
\forall \ell \in [0,1[, & \Pi_M(\ell) \in \CLOWER.
\end{align*}

We can then build a path $\Pi$ which is the concatenation of $\Pi_m$ and $\ell \rightarrow \Pi_M(1-\ell)$, which goes from $\xmin$ to $\XLOWER$ and goes through $\xsad$. Since this path stays inside $\CSAD$, we know that $\effort(\Pi) \leq f(\xsad) - f(\xmin)$, and then $\dynamics(\xmin) \leq f(\xsad) - f(\xmin)$.

\medskip

By grouping the two inequalities, we obtain that $\dynamics(\xmin) = f(\xsad) - f(\xmin)$, and then by uniqueness of the critical values of $f$, $\xmin$ is then paired by dynamics to $\xsad$. \qed

\subsection{Pairing by dynamics implies pairing by persistence in $n$-D}

\begin{Th}
Let $f$ be a Morse function from $\Reals^n$ to $\Reals$. We assume that the local minimum $\xmin$ of $f$ is paired by dynamics to a $1$-saddle of $f$ of abscissa $\xsad$. Then, $\xsad$ is paired by persistence to $\xmin$.
\end{Th}

\ProofNico: Let us assume that $\xmin$ is paired to $\xsad$ by dynamics. Let us recall the usual framework relative to persistence:
\begin{align*}
\CSAD &= \CC([f \leq f(\xsad)],\xsad),\\
\{\CI\}_{i \in I} &= \CoCo([f < f(\xsad)]),\\
\{\CSADI\}_{i \in \ISAD} &= \left\{ \CI | \xsad \in \CLO(\CI) \right\},\\
\forall i \in \ISAD, \ \rep_i &= {\arg \min}_{x \in \CSADI} f(x).
\end{align*}

By Definition~\ref{def.persistence}, $\xsad$ is  paired to the representative $\rep_i$ of $\CSADI$ which maximizes $f(\rep_i)$. 

\medskip

\begin{enumerate}

\item Let us show that there exists some index $\imin$ such that $\xmin$ is the representative of a component $\CSADIMIN$ of $\{\CSADI\}_{i \in \ISAD}$.

\begin{enumerate}

    \item First, $\xmin$ is paired by dynamics with $\xsad$ and $\dynamics(\xmin)$ is greater than zero, then $f(\xsad) > f(\xmin)$, then $\xmin$ belongs to $[f < f(\xsad)]$, then there exists some $\imin \in I$ such that $\xmin \in \CIMIN$ (see Equation $(2)$ above).

    \item Now, if we assume that $\xmin$ is not the representative of $\CIMIN$, there exists then some $\XLOWER$ in $\CIMIN$ satisfying that $f(\XLOWER) < f(\xmin)$, and then there exists some $\Pi$ in $\DXMIN$ whose image is contained in $\CIMIN$. In other words,
    $$\dynamics(\xmin) \leq \effort(\Pi) < f(\xsad) - f(\xmin),$$
    which contradicts the hypothesis that $\xmin$ is paired with $\xsad$ by dynamics.

    \item Let us show that $\imin$ belongs to $\ISAD$, that is, $\xsad \in \CLO(\CIMIN)$. Let us assume that:
    $$\xsad \not \in \CLO(\CIMIN). \ \ \ (\HYPDEUX)$$
    Every path in $\DXMIN$ goes outside of $\CIMIN$ to reach some point whose image by $f$ is lower than $f(\xmin)$ since $\xmin$ has been proven to be the representative of $\CIMIN$. Then this path  intersects the boundary $\bd$ of $\CIMIN$. Since by $(\HYPDEUX)$, $\xsad$ does not belong to the boundary $\bd$ of $\CIMIN$, any optimal path $\PISTAR$ in $\DXMIN$  goes through one $1$-saddle $\xsad_2 = {\arg \max}_{\ell \in [0,1]} f(\PISTAR(\ell))$ (by Lemma~\ref{lemma.saddle}) different from $\xsad$ and \nico{satisfying} then $f(\xsad_2) > f(\xsad)$. Thus, $\dynamics(\xmin) > f(\xsad) - f(\xmin)$, which contradicts the hypothesis that $\xmin$ is paired with $\xsad$ by dynamics. Then, we have:
    $$\xsad \in \CLO(\CIMIN).$$

\end{enumerate}

\item Now let us show that $f(\xmin) > f(\rep(\CSADI))$ for any $i \in \ISAD \setminus \{\imin\}$. For this aim, we prove that there exists some $i \in \ISAD$ such that $f(\rep(\CSADI)) < f(\xmin)$ and we  conclude with Proposition~\ref{proposition.2}. Let us assume that the  representative $r$ of each component $\CSADI$ except $\CMIN$ satisfies $f(r) > f(\xmin)$, then any path $\Pi$ of $\DXMIN$  has to go outside $\CSAD$ to reach some point whose image by $f$ is lower than $f(\xmin)$. We obtain the same situation as before (see $(1.c)$), and then we obtain that the effort of $\Pi$ is greater than $f(\xsad) - f(\xmin)$, which leads to a contradiction with the hypothesis that $\xmin$ is paired with $\xsad$ by dynamics. We have then that there exists $i \in \ISAD$ such that $f(\rep(\CSADI)) < f(\xmin)$. Thanks to Proposition~\ref{proposition.2}, we know then that $\xmin$ is the representative of the components of $[f < f(\xsad)]$ whose image by $f$ is the greatest.

\item It follows that $\xsad$ is paired with $\xmin$ by persistence.

\end{enumerate}

\section{Perspectives: a research program linking Topological Data Analysis and MM}
\label{sec:perspectives}

\begin{figure*}
\centering
\includegraphics[width=\linewidth]{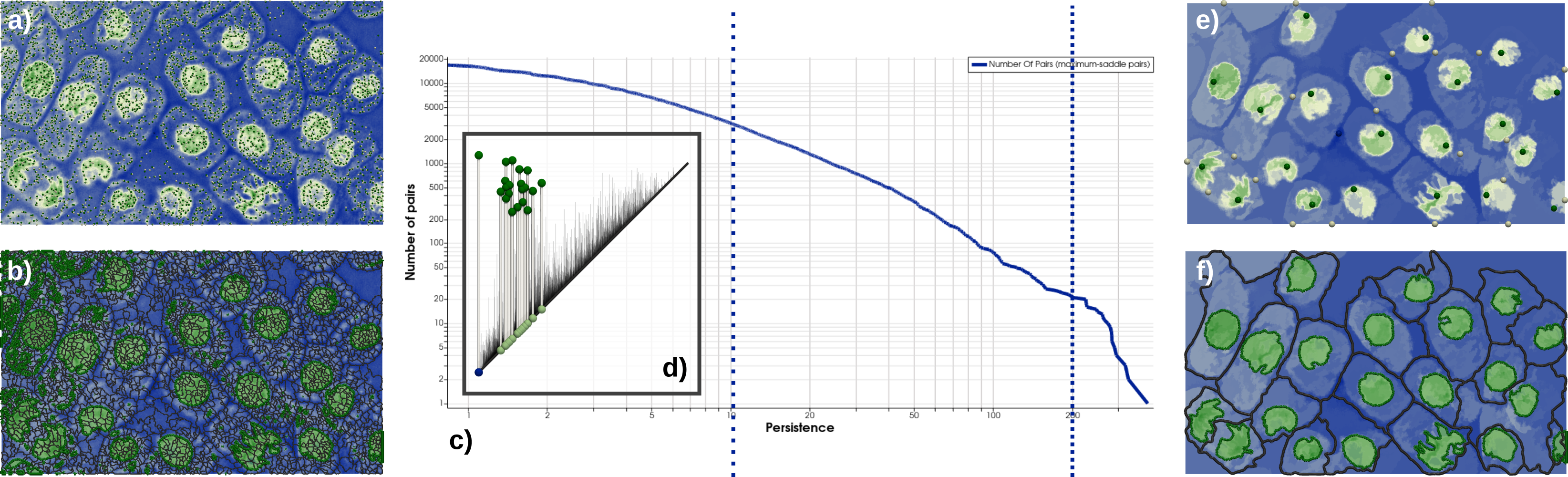}
\caption{An example of segmentation of a microscopy image of cells and their nuclei~\cite{lukasczyk2020localized} with  the topological data analysis framework. The very same example can be seen  as an application of the morphological data analysis framework (see text).}
\label{fig.TDApipeline}
\end{figure*}

\begin{figure*}
    \centering
    \includegraphics[width=.7\linewidth]{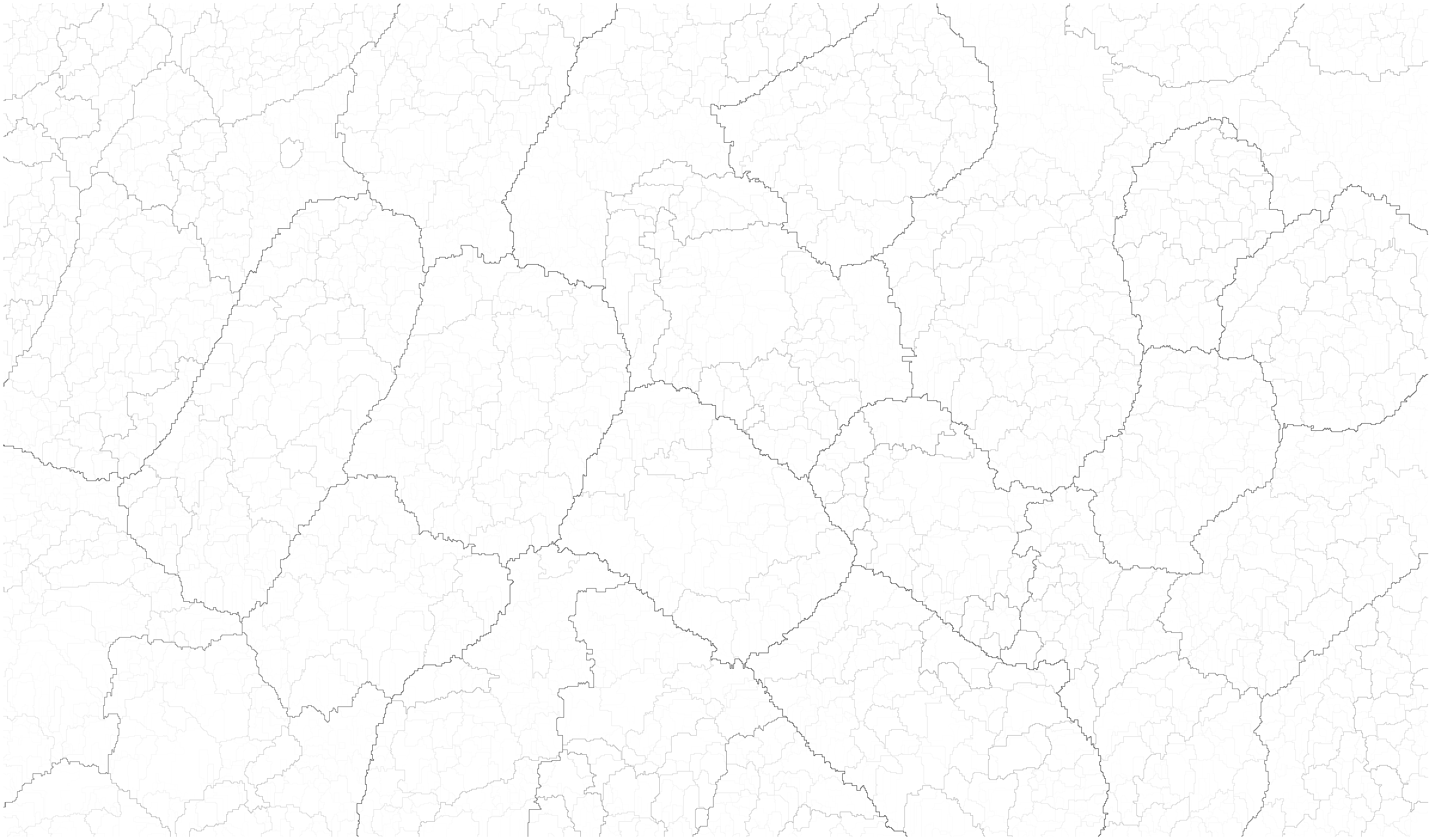}
    \caption{Saliency map corresponding to Fig.\ref{fig.TDApipeline}. In this image, the contours that are the more persistent are darker than the others (see text for details.)}
    \label{fig:saliencyCells}
\end{figure*}

This paper is a step towards exploring the possible interactions between Topological Data Analysis (TDA) and MM. In this section, we detail some ideas for a research program linking these two fields. 

As a very first example, let us look at
Fig.\ref{fig.TDApipeline}, which provides an illustration of an image analysis pipe\-line originally performed in the context of topological data analysis using the library \nico{called \emph{Topology Toolkit}~\cite{tierny2017topology,masood2019overview} (shortly TTK).}
In the original publication \cite{lukasczyk2020localized}, the steps are the following
\begin{enumerate}
    \item The original data (microscopy image of cells and their nuclei) are simplified with a small threshold of persistence (Fig.~\ref{fig.TDApipeline}.a)
    \item The Morse-Smale complex leads to an oversegmentation ((Fig.~\ref{fig.TDApipeline}.b)
    \item The persistence curve (Fig.~\ref{fig.TDApipeline}.c) is the number of persistent pairs as a function of their persistence. The vertical dashed line on the left corresponds to the level of simplification of Fig.~\ref{fig.TDApipeline}.a and b. The vertical dashed line on the right corresponds to the level of simplification of Fig.~\ref{fig.TDApipeline}.e and f.
    \item The diagram of persistence (Fig.~\ref{fig.TDApipeline}.d) 
    \item The image is simplified (Fig.~\ref{fig.TDApipeline}.e) with a threshold corresponding to the vertical dashed line on the right of Fig.~\ref{fig.TDApipeline}.c.
    \item The Morse-Smale complex separatrices of Fig.~\ref{fig.TDApipeline}.e provides 1 maximum per nuclei, while the nuclei are the maxima of the same image (Fig.~\ref{fig.TDApipeline}.f).
\end{enumerate}

Thanks to the result of this paper and some previous work, we can translate this process in mathematical morphology. The filtering by persistence belong to a class of morphological filters called {\em connected filters} \cite{salembier2009connected}, with a criterion named dynamics. The Morse-Smale complexe is replaced by the watershed \cite{vcomic2005morse,vcomic2016computing}. The persistence curve is called a granulometric curve \cite{matheron1972random}. Hence, from a morphological perspective, the same example can be done using Higra \cite{perret2019higra}, a (morphological) library that computes the various steps, and this leads to the following description.
\begin{enumerate}
    \item A connected filter with a small dynamics threshold is first applied on the original data (Fig.~\ref{fig.TDApipeline}.a)
    \item The watershed of Fig.~\ref{fig.TDApipeline}.a is oversegmented (see Fig.~\ref{fig.TDApipeline}.b)
    \item The granulometric curve (Fig.~\ref{fig.TDApipeline}.c) provides the number of maximum as a function of the dynamics
    \item A connected filter of Fig.~\ref{fig.TDApipeline}.a  with a dynamics thres\-hold corresponding to the vertical dashed line on the right of Fig.~\ref{fig.TDApipeline}.c leads to Fig.~\ref{fig.TDApipeline}.e.
    \item The watershed of Fig.~\ref{fig.TDApipeline}.e gives one region per cell, while the nuclei are the maxima of the same image (Fig.~\ref{fig.TDApipeline}.f).
\end{enumerate}

It is worthwhile to explore the differences between the two approaches. In mathematical morphology, there is no persistence diagram. On the other hand, there exist saliency maps \cite{najman1996geodesic,najman2011equivalence,cousty2018hierarchical}. Intuitively, a saliency map can be obtained by filtering the original image/data for all values of the criterion (here, dynamics), and stacking (summing) the watersheds of all the filtered images. A contour that is persistent is present many times in the stack, and has a high value in the resulting saliency map. Fig.~\ref{fig:saliencyCells} shows the saliency map of the original data of Fig.~\ref{fig.TDApipeline} for the dynamics criterion. 

In TDA, only a few criteria other than dynamics have been studied \cite{carr2004simplifying} but MM has many more, see \cite{Higra::criteria} for a few of them. There exist also several ways to simplify using non-increasing criteria \cite{salembier1998antiextensive,urbach2007connected,xu2015connected}.

The links between Morse-Smale Complex and watershed \cite{vcomic2005morse,vcomic2016computing} need to be explored, specifically in the context of Discrete Morse Theory~\cite{forman1998morse}. We envision doing such a study based on watershed cuts \cite{cousty2009watershed}, see also \cite{cousty2014collapses} that highlights some links between the watershed and topology. 

Many other comparisons should be done. To mention one of those, the {\em contour tree} \cite{freeman1967searching} from TDA is closely related to the {\em tree of shapes} \cite{caselles2009geometric} from MM. Comparing those trees and the algorithms for computing them from TDA \cite{carr2003computing,gueunet2019task} and from MM \cite{geraud2013quasi,crozet2014first,carlinet2018tree}  would be rewarding. In particular, the morphological algorithms for computing the tree of shapes, which are quasi-linear whatever the dimension of the space, are based on the ones for computing the tree of upper or lower level sets, called the {\em component trees} \cite{carlinet2014comparative}, and seem more efficient than the ones from TDA.

\section{Conclusion}
\label{sec.conclusion}

In this paper, we have proved that persistence and dynamics lead to the same pairings in $n$-D, $n \geq 1$, which implies that they are equal whatever the dimension. Concerning the future works, we propose to investigate the relationship between persistence and dynamics in the discrete case~\cite{forman1998morse} (that is, on complexes). We will also check under which conditions pairings by persistence and by dynamics are equivalent for functions that are not Morse. Furthermore, we will examine if the fast algorithms used in MM like watershed cuts, Betti numbers computations or attribute-based filtering are applicable to PH. Conversely, we will study if some PH concepts can be seen as the generalization of some MM concepts (for example, dynamics seems to be a particular case of persistence).

More generally, we believe that exploring the links and differences between TDA and MM would benefit to the two communities.

\section*{Acknowledgements}
The authors would like to thank Julien Tierny for many interesting discussions and for providing \nico{us} Fig. \ref{fig.TDApipeline}.

\appendix
\section{Ambiguities occurring when values are not unique}
\label{app:ambiguities}

\begin{figure}[h!]
    \centering
    \includegraphics[width=0.5\linewidth]{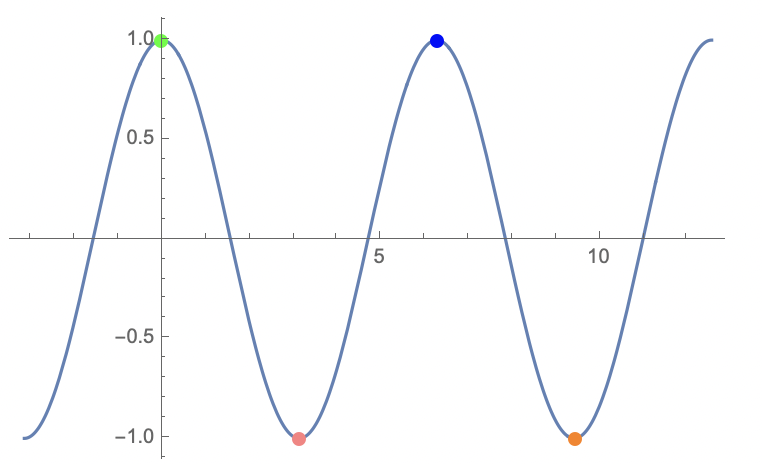}
    \caption{Ambiguities can occur when critical values are not unique for pairing by dynamics and for pairing by persistence.}
    \label{fig:ambiguities}
\end{figure}

As depicted in Figure~\ref{fig:ambiguities}, the abscissa of the blue point can be paired by persistence to the abscissas of the orange and/or the red points. The same thing appears when we want to pair the abscissa of the pink point to the abscissas of the green and/or blue points. This shows how much it is important to have unique critical values on Morse functions. This point is discussed in detail in \cite{bertrand2007dynamics}, where it is shown that a strict total order relation on the set of minima allows for a good definition of the dynamics.

\bibliographystyle{plain}
\bibliography{article}{}

\end{document}